\documentclass[journal]{IEEEtran}
\usepackage{times} 
\usepackage{amsmath,amsfonts}
\usepackage{amssymb}
\usepackage[ruled, vlined, linesnumbered]{algorithm2e}
\usepackage{array}
\usepackage[caption=false,font=normalsize,labelfont=sf,textfont=sf]{subfig}
\usepackage{textcomp}
\usepackage{stfloats}
\usepackage{url}
\usepackage{verbatim}
\usepackage{graphicx}
\usepackage{cite}
\usepackage{multirow} 
\usepackage{xcolor}
\definecolor{myblue}{RGB}{91, 155, 213}
\begin{document}

\title{SemDiff: Generating Natural Unrestricted Adversarial Examples via Semantic Attributes Optimization in Diffusion Models}

\author{Zeyu Dai, Shengcai Liu, Rui He, Jiahao Wu, Ning Lu, Wenqi Fan, Qing Li,~\IEEEmembership{Fellow,~IEEE,} and Ke Tang,~\IEEEmembership{Fellow,~IEEE}
\thanks{Manuscript received April 6, 2025.}
\thanks{Zeyu Dai, Shengcai Liu, Rui He, Jiahao Wu, Ning Lu and Ke Tang are with the Guangdong Provincial Key Laboratory of Brain-Inspired Intelligent Computation and Department of Computer Science and Engineering, Southern University of Science and
Technology, Shenzhen 518055, China (e-mail: liusc3@sustech.edu.cn).}
\thanks{Zeyu Dai, Jiahao Wu, Wenqi Fan and Qing Li are with the Department of Computing, The Hong Kong Polytechnic University, Hong Kong.}
}

\markboth{Journal of \LaTeX\ Class Files,~Vol.~14, No.~8, August~2021}%
{Dai \MakeLowercase{\textit{et al.}}: SemDiff: Generating Semantic Unrestricted Adversarial Examples with Diffusion Models}

\IEEEpubid{0000--0000/00\$00.00~\copyright~2021 IEEE}

\maketitle

\begin{abstract}
Unrestricted adversarial examples (UAEs), allow the attacker to create non-constrained adversarial examples without given clean samples, posing a severe threat to the safety of deep learning models. 
Recent works utilize diffusion models to generate UAEs.
However, these UAEs often lack naturalness and imperceptibility due to simply optimizing in intermediate latent noises. 
In light of this, we propose SemDiff, a novel unrestricted adversarial attack that explores the semantic latent space of diffusion models for meaningful attributes, and devises a multi-attributes optimization approach to ensure attack success while maintaining the naturalness and imperceptibility of generated UAEs. 
We perform extensive experiments on four tasks on three high-resolution datasets, including CelebA-HQ, AFHQ and ImageNet. 
The results demonstrate that SemDiff outperforms state-of-the-art methods in terms of attack success rate and imperceptibility. The generated UAEs are natural and exhibit semantically meaningful changes, in accord with the attributes' weights. 
In addition, SemDiff is found capable of evading different defenses, which further validates its effectiveness and threatening. 
\end{abstract}

\begin{IEEEkeywords}
Unrestricted adversarial examples, adversarial attacks, diffusion models, semantic latent space, imperceptibility.
\end{IEEEkeywords}

\section{Introduction}
\IEEEPARstart{D}{eep} Neural Networks (DNNs) have achieved significant success in wide range of applications, such as image classification~\cite{1}, face recognition~\cite{2}, social recommendation~\cite{3}, and machine translation~\cite{4}. 
However, a lot of research finds that deep learning models are vulnerable to adversarial attacks~\cite{5,6,7,8,9,10,11,12,13,14,15,16,17,18}. 
Traditional adversarial attacks try to generate adversarial examples to fool DNNs into wrong predictions by injecting imperceptible perturbations into clean samples~\cite{5,6,7,8}. 
This type of adversarial examples is called as perturbation-based adversarial examples~\cite{12}, which pose a constraint on the perturbation magnitude~\cite{6,7,8}.

Different from perturbation-based adversarial examples, unrestricted adversarial examples (UAEs) have no common perturbation constraints and do not depend on any given clean images, hence are more threatening to the safety of deep learning models~\cite{12,13,14,15,16,17,18}.
Specifically, common unrestricted adversarial attacks use generative models to generate UAEs by optimizing the random noise (e.g., the input noise vector of Generative Adversarial Networks (GANs)~\cite{12,13,14} or the intermediate latent noise of diffusion models~\cite{17,18}).
Since the random noise could refer to unseen samples in the dataset and there is no limitation in the $L_p$ norm spaces from existing data samples, UAEs are more flexible and aggressive than perturbation-based adversarial examples~\cite{16}.
Moreover, attackers have no need to collect clean samples as the inputs in advance, and can generate an infinite number of UAEs with a trained generative model~\cite{17}.
This is particularly important in some data-expensive attack scenarios, such as medical images~\cite{19}, where collecting clean samples is difficult and expensive.

\IEEEpubidadjcol

\begin{figure}[!t]
    \centering
    \includegraphics[width=3in]{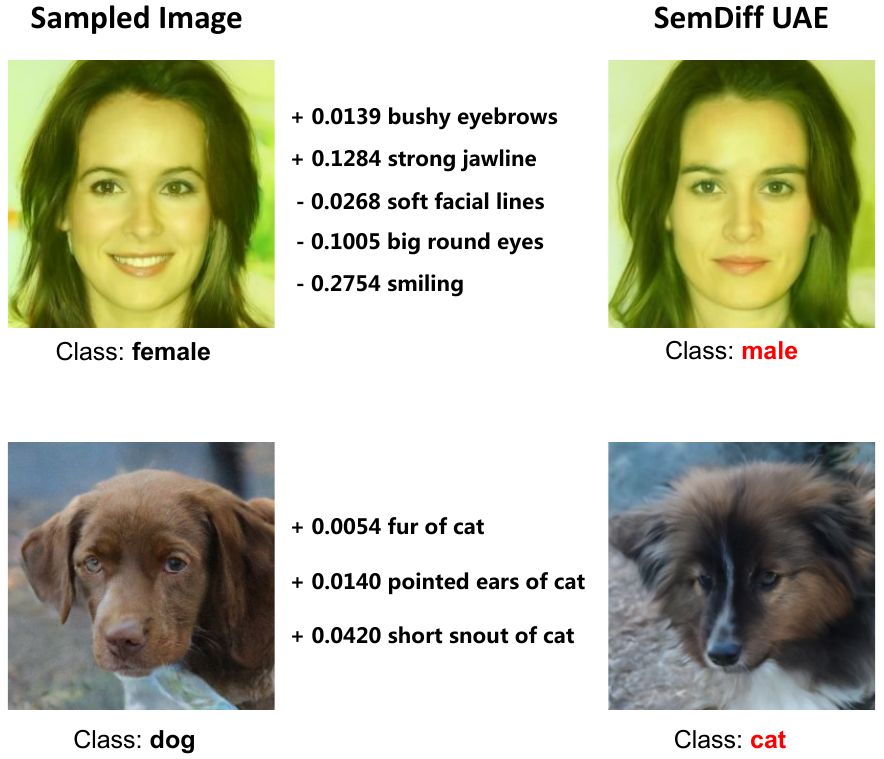}
    \caption{\textbf{Some UAEs generated by the proposed SemDiff.} Note that the sampled images are not real-world images, but are generated by diffusion models with randomly sampled noise images. Our SemDiff optimizes the weights of multiple meaningful attributes to craft adversarial examples based on the sampled images. It is shown that the weights are in accord with the semantic changes in images, even the negative weight can lead to an opposite but meaningful change.}
    \label{fig1}
\end{figure}

Previous UAE works utilize GANs to generate UAEs~\cite{12,13,14,15,16}, but they show unsatisfactory visual quality on high-resolution datasets due to GAN's limited interpretability and unstable training~\cite{17}. 
Instead, recent works~\cite{17,18} adopt diffusion models, which outperform GAN-based models in terms of image synthesis quality and diversity~\cite{20,21,22}.
However, the generated UAEs of these works are still visually unnatural, exhibiting local visible perturbations (see the examples of AdvDiff~\cite{17} and AdvDiffuser~\cite{18} in Fig.~\ref{fig3} and Fig.~\ref{fig4}).
The root cause is that they directly optimize in the intermediate latent noise space of diffusion models~\cite{17,18}.
Unlike GANs, whose input latent space encodes high-level semantic information, the intermediate latent noise of diffusion models are dominated by stochastic noise~\cite{30} and hard to directly capture such semantics~\cite{23,24}.
This noise-centric nature forces perturbations to focus on low-level pixel variations rather than semantic attributes.
Furthermore, diffusion models are found to construct the high-level context in the early stage and refine details in the later stage of the denoising process~\cite{25}.
However, because the intermediate latent images in the early steps are too noisy to obtain a accurate gradient, these works~\cite{17,18} primarily perturb in the later steps, which only affect local details.
As a result, the generated UAEs fail to achieve semantic coherence and exhibit unnatural local patterns similar to traditional perturbation-based adversarial examples.

In order to address these issues, we propose \textbf{SemDiff}, a novel unrestricted adversarial attack that explores the semantic latent space of diffusion models to generate natural and realistic UAEs with semantic attribute changes.
Specifically, leveraging the disentanglement property of diffusion models~\cite{24}, we propose to learn some meaningful but adversarial attributes in the semantic latent space of diffusion models.
The semantic latent space lies in the deepest feature maps in the UNet of diffusion models.
In contrast to the intermediate latent noise, these bottleneck features encode high-level semantics with minimal spatial redundancy~\cite{24}.
Then, we create a set of meaningful and class-sensitive attributes for each task, such as \textit{bushy eyebrows} and \textit{strong jawline} for gender classification task on CelebA-HQ dataset.
Changing the weight of such attribute in the semantic latent space can lead to the corresponding attribute change in images.

Furthermore, considering a single attribute needs a large weight to successfully fool target models that often results in excessive semantic shifts, we devise a multi-attributes optimization approach.
The weights of multiple attributes are jointly optimized together with a weight penalty term, hence slight changes of multiple attributes are sufficient to cause misclassification while keeping UAEs realistic and close to the sampled images, as exemplified in Fig.~\ref{fig1}.
Compared with current diffusion-based UAE works~\cite{17,18}, our SemDiff updates images along the direction of semantic attributes, generating more natural UAEs with imperceptible difference from real-world images.

We summarize our contributions as follows:
\begin{itemize}
    \item To the best of our knowledge, our work is the first to utilize the semantic latent space of diffusion models to generate UAEs. Previous works only optimize in the intermediate latent noise space, which hinders the naturalness and imperceptibility of UAEs.
    
    \item We propose SemDiff, a novel unrestricted adversarial attack that optimizes the weights of multiple meaningful attributes in the semantic latent space of diffusion models, achieving attack success while maintaining the naturalness of generated UAEs.
    
    \item We conduct extensive experiments across four tasks on three high-resolution datasets (CelebA-HQ, AFHQ and ImageNet). The results demonstrate that SemDiff outperforms state-of-the-art methods in terms of attack success rate and imperceptibility. The generated UAEs are more natural with semantic attribute changes.
    
    \item We empirically show that SemDiff can bypass various defenses (detection-based defenses, preprocessing-based defenses and adversarial training), which further validates the threatening of our attack, and unveils the vulnerability of current deep neural networks and defenses.
\end{itemize}

\section{Related Work and Preliminary}
\noindent In this section, we first review the traditional perturbation-based adversarial examples. Then we discuss the related work on unrestricted adversarial examples, including GAN-based UAEs and diffusion-based UAEs. Finally, we introduce diffusion models and their disentanglement capability.

\subsection{Perturbation-based Adversarial Examples}
Early adversarial attack works focus on perturbation-based adversarial examples, which add a small perturbation into clean samples to fool the target models into incorrect predictions~\cite{5,6,7,8,26,27,53,54}.
The perturbation is usually constrained in a $L_p$ norm space to keep imperceptible to human eyes.
Perturbation-based adversarial examples can be defined as~\cite{12}:
\begin{multline}
    \mathcal{A}_{p} \triangleq \{\tilde{\boldsymbol{x}} \in \mathcal{O} \mid \exists \boldsymbol{x} \in \mathcal{T}, ||\tilde{\boldsymbol{x}} - \boldsymbol{x}||_{p} \leq \epsilon \\
    \wedge \boldsymbol{f}(\boldsymbol{x}) = \boldsymbol{o}(\boldsymbol{x}) = \boldsymbol{o}(\tilde{\boldsymbol{x}}) \neq \boldsymbol{f}(\tilde{\boldsymbol{x}})\},
\end{multline}
where $\mathcal{O}$ is the set of all images that look realistic to humans, $\mathcal{T}$ is a selected test dataset $\mathcal{T} \subset \mathcal{O}$, $|| \cdot ||_{p}$ is a $L_p$ norm, such as $L_0$, $L_2$ or $L_{\infty}$. 
$\epsilon$ is a small constraint, $\boldsymbol{f}$ is the target model to be attacked, and $\boldsymbol{o}$ represents an oracle model that provides ground-truth results.
A lot of methods have been proposed to generate perturbation-based adversarial examples, such as the fast gradient-sign method (FGSM)~\cite{5} with a $L_{\infty}$ norm constraint, projected gradient descent (PGD)~\cite{26} with a $L_{2}$ norm constraint, and $L_{0}$ norm Jacobian-based Saliency Map Attack (JSMA)~\cite{27}.

\subsection{Unrestricted Adversarial Examples}
Song et al.~\cite{12} first proposed the concept of unrestricted adversarial examples (UAEs), which get rid of the dependence of input images, as long as the generated UAEs do not change the semantics. The definition of UAEs can be formulated as:
\begin{equation}
    \mathcal{A}_{u} \triangleq \{\tilde{\boldsymbol{x}} \in \mathcal{O} \mid \boldsymbol{o}(\tilde{\boldsymbol{x}}) \neq \boldsymbol{f}(\tilde{\boldsymbol{x}})\}.
\end{equation}
It is clear that the set of UAEs is a superset of that of traditional perturbation-based adversarial examples, namely $\mathcal{A}_{p} \subset \mathcal{A}_{u}$. To craft UAEs, current researchers utilize generative models, including GANs and diffusion models.

\subsubsection{GAN-based UAEs}
One line of works first build a generator of GANs to generate realistic images, and then optimize the input noise vector to update the output to fool the target model~\cite{12,13,14}.
Specifically, Song et al.~\cite{12} proposed to use a well-trained AC-GAN~\cite{28} to generate UAEs.
Starting from a randomly sampled noise vector $z^0$, the method searches a $z$ by minimizing an adversarial loss to achieve misclassification.
In addition, Khoshpasand and Ghorbani~\cite{13} trained a generator for each class to improve the transferability of UAEs.
Zhang et al.~\cite{14} designed a novel augmented triple-GAN and train a MLP to produce the noise vector instead of optimizing it directly.
Another line of GAN-based UAEs is end-to-end method, which trains generators to directly output realistic UAEs that can deceive the target model~\cite{15,16} without the optimization of noise vector.
However, all of these methods only focus on small datasets, and their UAEs are found low-quality and unnatural on high-resolution datasets such as ImageNet~\cite{17}.

\subsubsection{Diffusion-based UAEs}
Inspired by recent advances of diffusion models in image synthesis, some recent works turn to utilize diffusion models to generate UAEs~\cite{17,18}.
Chen et al.~\cite{18} first proposed AdvDiffuser to generate UAEs with a pre-trained conditional DDPM~\cite{20}.
Starting from an initial random noise $x_T$, AdvDiffuser performs PGD to add perturbation into each intermediate latent noise in the reverse process, and obtains the final output $x_0$ as the UAE.
Furthermore, Dai et al.~\cite{17} proposed AdvDiff and added another noise sampling guidance to update the initial noise $x_T$ in each iteration, achieving a higher attack success rate.
Although theses diffusion-based methods achieve better performance than GAN-based methods, we find that the generated UAEs are still unnatural with visible and meaningless perturbations as they directly perturb in the intermediate latent noises of diffusion models.
In contrast, our proposed SemDiff explores the semantic latent space of diffusion models to generate UAEs with natural and meaningful attribute changes.

\subsection{Diffusion Models and Disentanglement}
\label{sec:diffusion}
Dinh et al.~\cite{29} first proposed the concept of diffusion-based generation. 
After that, Ho et al.~\cite{30} proposed denoising diffusion probabilistic models (DDPMs), which significantly improved both the quality and diversity of the generated images, making diffusion models a competitive alternative to traditional GANs.
Specifically, the forward process of DDPMs starts from an real image $\boldsymbol{x}_0$ and progressively diffuses it in a sequence of steps $\boldsymbol{x}_{1:T}$ through Gaussian transitions, which is defined as a Makov chain:
\begin{equation}
    q\left(\boldsymbol{x}_t \mid \boldsymbol{x}_{t-1}\right)=\mathcal{N}\left(\boldsymbol{x}_t ; \sqrt{1-\beta_t} \boldsymbol{x}_{t-1}, \beta_t \mathbf{I}\right),
\end{equation}
where $\{\beta\}^{T}_{t=1}$ is the variance schedule. 
As $T \rightarrow \infty$, $\boldsymbol{x}_T$ approaches an isotropic Gaussian distribution.
Then the reverse process gradually denoises the noise and finally recovers the real image:
\begin{equation}
    p_\theta\left(\boldsymbol{x}_{t-1} \mid \boldsymbol{x}_t\right)=\mathcal{N}\left(\boldsymbol{x}_{t-1} ; \boldsymbol{\mu}_\theta\left(\boldsymbol{x}_t, t\right), \sigma^{2}_{t} \mathbf{I} \right),
\end{equation}
where $\sigma^{2}_{t}$ is a variance of the reverse process and is set to $\beta_t$. The mean $\boldsymbol{\mu}_\theta$ can be modeled with a noise predictor $\boldsymbol{\epsilon}^{\theta}_t$:
\begin{equation}
    \boldsymbol{x}_{t-1}=\frac{1}{\sqrt{1-\beta_t}}\left(\boldsymbol{x}_t-\frac{\beta_t}{\sqrt{1-\alpha_t}} \boldsymbol{\epsilon}_t^\theta\left(\boldsymbol{x}_t\right)\right)+\sigma_t \boldsymbol{z}_t,
\end{equation}
where $\alpha_t=\prod_{i=1}^t\left(1-\beta_i\right)$ and $\boldsymbol{z}_t \sim \mathcal{N}\left(0, \mathbf{I}\right)$. 
Subsequently, DDPMs learn the data distribution by training $\boldsymbol{\epsilon}_t^\theta$ using the variational lower-bound (VLB).

In addition, Song et al.~\cite{31} proposed denoising diffusion implicit model (DDIM), which defines a non-Markovian process and achieves much faster sampling speed with fewer steps:
\begin{equation}\label{eq6}
    \boldsymbol{x}_{t-1}=\sqrt{\alpha_{t-1}} \ \mathbf{P}_{t}(\boldsymbol{\epsilon}_t^\theta\left(\boldsymbol{x}_t\right)) + \mathbf{D}_{t}(\boldsymbol{\epsilon}_t^\theta\left(\boldsymbol{x}_t\right)) + \sigma_t \boldsymbol{z}_t,
\end{equation}
where $\mathbf{P}_{t}(\boldsymbol{\epsilon}_t^\theta\left(\boldsymbol{x}_t\right)) = \left(\frac{\boldsymbol{x}_t-\sqrt{1-\alpha_t} \boldsymbol{\epsilon}_t^\theta\left(\boldsymbol{x}_t\right)}{\sqrt{\alpha_t}}\right)$ denotes the predicted $\boldsymbol{x}_0$, $\mathbf{D}_{t}(\boldsymbol{\epsilon}_t^\theta\left(\boldsymbol{x}_t\right)) = \sqrt{1-\alpha_{t-1}-\sigma_t^2} \cdot \boldsymbol{\epsilon}_t^\theta\left(\boldsymbol{x}_t\right)$ denotes the direction pointing to $\boldsymbol{x}_t$, and $\sigma_t \boldsymbol{z}_t$ is a random noise.
We further abbreviate $\mathbf{P}_{t}(\boldsymbol{\epsilon}_t^\theta\left(\boldsymbol{x}_t\right))$ as $\mathbf{P}_{t}$ and $\mathbf{D}_{t}(\boldsymbol{\epsilon}_t^\theta\left(\boldsymbol{x}_t\right))$ as $\mathbf{D}_{t}$ for simplicity when the context clearly specifies the arguments.
$\sigma_t=\eta \sqrt{\left(1-\alpha_{t-1}\right) /\left(1-\alpha_t\right)} \sqrt{1-\alpha_t / \alpha_{t-1}}$, where $\eta$ is a hyperparameter that controls the variance of the noise, the process becomes DDPM when $\eta=1$ and DDIM when $\eta=0$.

Disentanglement capability allows image generative models to disentangle different attributes of the generated images, such as semantic contents and styles, which is crucial for image editing tasks.
Different from GANs inherently endowed with strong disentangled semantics in their latent space, diffusion models show degraded performance by simply editing in the intermediate latent noise space~\cite{23}.
Recent works have explored ways to achieve disentanglement in diffusion models. 
For instance, Preechakul et al.~\cite{32} introduced an additional input to the reverse sampling process.
The input is a semantic latent vector from an extra encoder, that is jointly trained with diffusion models.
Wu et al.~\cite{33} adopted stable diffusion~\cite{34} to achieve disentanglement by optimizing the mixing weights of two text embeddings.
In addition, Kwon et al.~\cite{24} proposed to leverage the deepest feature maps in UNet as the semantic latent space to edit images.
\section{Methodology}
\noindent In this section, we first introduce the motivations and overview of the proposed method. Then, we present the proposed adversarial semantic attributes and the multi-attributes optimization approach in detail.

\subsection{Motivations}
\begin{figure*}[htbp]
    \centering
    \includegraphics[width=\textwidth]{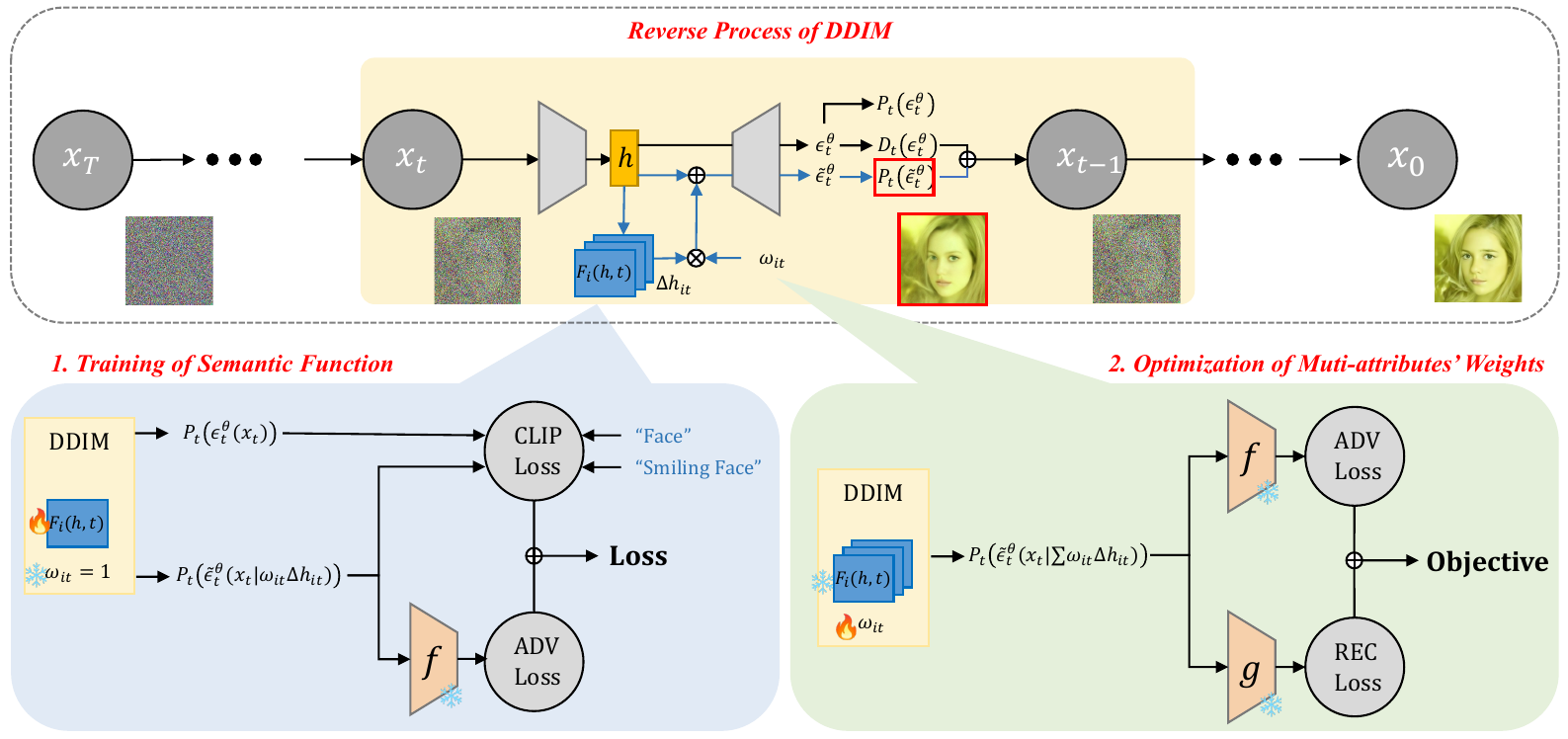}
    \caption{\textbf{Overview of the proposed SemDiff.} The yellow box illustrates that SemDiff only modifies $\mathbf{P}_{t}$ ({\color{myblue} blue} lines) while preserving $\mathbf{D}_{t}$ (black lines) in the reverse process of DDIM to alter the semantics of the generated UAEs. SemDiff first train a semantic function $\mathbf{F}_{i}(\boldsymbol{h}_{t}, t)$ to learn each adversarial semantic attribute with the loss shown in the blue box. Then, the weights of multiple attributes $\boldsymbol{w}_{it}$ are optimized with the objective exhibited in the green box. $\boldsymbol{f}$ is the target classifier to be attacked, $\boldsymbol{g}$ is another auxiliary classifier to maintain the class appearance. Notice that we use clearer $\mathbf{P}_{t}$ (within the {\color{red} red} border) instead of $\boldsymbol{x}_t$ at each timestep in both function training and weight optimization.}
    \label{fig2}
\end{figure*}

We aim to design an unrestricted adversarial attack that can generate natural and realistic adversarial examples to deceive target models without given clean samples.
Current state-of-the-art methods~\cite{17,18} utilize diffusion models to generate UAEs, which are still visually unnatural.
We find two issues that may degrade the naturalness of generated UAEs.
First, these diffusion-based methods directly add perturbations in the intermediate latent noise space, that is noisy and lack high-level semantic information like the input latent space of GANs.
Therefore, compared with GAN-based methods and our proposed SemDiff, existing diffusion-based methods fail to alter semantic attributes instead introduce perceptible perturbations into sampled images (see Fig.~\ref{fig3} and Fig.~\ref{fig4}), which hinders the naturalness and imperceptibility performance.
Our method is motivated to explore a high-level semantic latent space within diffusion models to generate natural UAEs with meaningful attribute changes, achieving better imperceptibility than existing methods.

Second, these diffusion-based methods~\cite{17,18} restrict perturbations to the later stages of the reverse denoising process, as the early-stage intermediate latent images are deemed too noisy for stable gradient computation~\cite{17,18}.
This design choice fundamentally conflicts with the hierarchical nature of diffusion models—prior work~\cite{25} finds that high-level semantics are established in early denoising steps, while later steps refine low-level details.
By perturbing only the later stages, these methods limit their impact to superficial details rather than semantic content, causing generated UAEs to retain unnatural local perturbations similar to traditional perturbation-based adversarial examples (see Fig.~\ref{fig3} and Fig.~\ref{fig4}).
To avoid this issue, we propose to optimize in the semantic latent space of diffusion models, and utilize $\mathbf{P}_{t}$ of DDIM (see Sec.~\ref{sec:diffusion}) instead of the intermediate latent noise $\boldsymbol{x}_t$ to obtain the adversarial gradient.
Compared to $\boldsymbol{x}_t$, $\mathbf{P}_{t}$ denotes the predicted $\boldsymbol{x}_0$ and removes the noise term, thus better approximating the final output $\boldsymbol{x}_0$ and clearer in early stage (see Fig.~\ref{fig6} for example).
We use $\mathbf{P}_{t}$ both in the training of semantic function and the optimization of multi-attributes' weights, making editing image semantics in the early stage of the denoising process possible.

To accomplish our goal, we propose \textbf{SemDiff}, a novel unrestricted adversarial attack, with overall framework depicted in Fig.~\ref{fig2}.
In order to generate a natural UAE, our method starts from a randomly sampled noise $\boldsymbol{x}_T$ and iteratively denoises it until the final output $\boldsymbol{x}_0$, adding perturbations in the semantic latent space $\boldsymbol{h}$ of diffusion models.
The perturbation is composed of multiple $\boldsymbol{\Delta h}_{it}$ that refer to different adversarial semantic attributes $\boldsymbol{a}_i$ at timestep $t$, and their corresponding weights $\boldsymbol{w}_{it}$.
Therefore, we firstly design a semantic function to learn different semantic attributes but easily deceive target models.
Furthermore, we devise a multi-attributes optimization approach to optimize the weights of adversarial semantic attributes, so as to further boost the attack success rate while maintaining the naturalness of generated UAEs.

\subsection{Adversarial Semantic Attributes}
We first introduce adversarial semantic attributes that can generate natural UAEs with semantically meaningful attribute changes but can deceive target models.

\subsubsection{Semantic Image Editing with Diffusion Models}
For semantic latent manipulation of images $\boldsymbol{x}_0$ given a \textit{pretrained} and \textit{frozen} diffusion model, we choose to shift the predicted noise $\boldsymbol{\epsilon}_t^\theta\left(\boldsymbol{x}_t\right)$ instead of the intermediate latent noise $\boldsymbol{x}_t$ at each sampling step.
However, it is found that the intermediate changes in $\mathbf{P}_{t}$ and $\mathbf{D}_{t}$ will destruct each other causing the same $p_\theta(\boldsymbol{x}_{0:T})$~\cite{24}.
Therefore, we only modify $\mathbf{P}_{t}$ by shifting $\boldsymbol{\epsilon}_t^\theta(\boldsymbol{x}_t)$ to $\tilde{\boldsymbol{\epsilon}}_t^\theta(\boldsymbol{x}_t)$ while preserving $\mathbf{D}_{t}$:
\begin{equation}
    \boldsymbol{x}_{t-1}=\sqrt{\alpha_{t-1}} \ \mathbf{P}_{t}(\tilde{\boldsymbol{\epsilon}}_t^\theta\left(\boldsymbol{x}_t\right)) + \mathbf{D}_{t}(\boldsymbol{\epsilon}_t^\theta\left(\boldsymbol{x}_t\right)) + \sigma_t \boldsymbol{z}_t.
    \label{eq7}
\end{equation}
Considering that $\boldsymbol{\epsilon}_t^\theta$ is usually implemented as UNet in diffusion models, we follow \cite{24} to modify its deepest feature maps $\boldsymbol{h}_t$ to control $\boldsymbol{\epsilon}_t^\theta$, because $\boldsymbol{h}_t$ has smaller spatial resolutions and high-level semantics than $\boldsymbol{\epsilon}_t^\theta$.
The sampling equation accordingly becomes:
\begin{equation}
    \boldsymbol{x}_{t-1}=\sqrt{\alpha_{t-1}} \ \mathbf{P}_{t}(\boldsymbol{\epsilon}_t^\theta\left(\boldsymbol{x}_t | \Delta \boldsymbol{h}_{t}\right)) + \mathbf{D}_{t}(\boldsymbol{\epsilon}_t^\theta\left(\boldsymbol{x}_t\right)) + \sigma_t \boldsymbol{z}_t,
\end{equation}
where $\boldsymbol{\epsilon}_t^\theta\left(\boldsymbol{x}_t | \Delta \boldsymbol{h}_{t}\right)$ adds $\Delta \boldsymbol{h}_{t}$ to the original feature maps $\boldsymbol{h}_t$.
Since directly optimizing $\Delta \boldsymbol{h}_{t}$ on multiple timesteps is computationally expensive and needs carefully chosen hyperparamenters, we instead use a semantic function $\mathbf{F}(\boldsymbol{h}_{t}, t)$ to generate $\Delta \boldsymbol{h}_{t}$ for given $\boldsymbol{h}_{t}$ and $t$.
$\mathbf{F}$ is implemented as a small neural network with two 1x1 convolutions concatenated with timestep $t$. See Supplementary Material A for the details of $\boldsymbol{h}_{t}$ and $\mathbf{F}$.

\subsubsection{CLIP Guidance}
For training a semantic function $\mathbf{F}$ to learn a specific semantic attribute, we use directional CLIP loss~\cite{35} with a pre-trained image encoder $E_I$ and text encoder $E_T$ of CLIP~\cite{36}:
\begin{equation}
    \label{eq9}
    \begin{gathered}
        \mathcal{L}_{\text{direction}}\left(\boldsymbol{x}^{\text{edit}}, c^{\text{target}} ; \boldsymbol{x}^{\text{source}}, c^{\text{source}}\right) = 1-\frac{\Delta I \cdot \Delta T}{\|\Delta I\|\|\Delta T\|}, \\ 
        \Delta T = E_T\left(c^{\text{target}}\right)-E_T\left(c^{\text{source}}\right), \\ 
        \Delta I = E_I\left(\boldsymbol{x}^{\text{edit}}\right)-E_I\left(\boldsymbol{x}^{\text{source}}\right),
    \end{gathered}
    \raisetag{10pt} 
\end{equation}
where $\boldsymbol{x}^{\text {edit}}$ and $\boldsymbol{x}^{\text {source}}$ are the edited and source images, $c^{\text {target}}$ and $c^{\text {source}}$ are the target and source textual descriptions.
Compared to directly minimizing the cosine distance between the edited image $E_I\left(\boldsymbol{x}^{\text{edit}}\right)$ and the target textual description $E_T\left(c^{\text{target}}\right)$, the directional CLIP loss aligns the direction of image transformation with the direction of textual transformation, which preserves image structural consistency and enables more precise attribute editing.
In our work, we use the modified $\mathbf{P}_{t}^{\text{edit}} = \mathbf{P}_{t}(\tilde{\boldsymbol{\epsilon}}_t^\theta)$ and the original $\mathbf{P}_{t}^{\text{source}} = \mathbf{P}_{t}(\boldsymbol{\epsilon}_t^\theta)$ as the edited and source images, and the textual prompts such as `smiling face' and `face' for attribute \textit{smiling} as the target and source textual descriptions.
We also add a regularization term to keep $\mathbf{P}_{t}^{\text{edit}}$ close to $\mathbf{P}_{t}^{\text{source}}$ and the original $\mathbf{P}_{t}^{\text{source}}$:
\begin{multline}
    \mathcal{L}_{\text{CLIP}} = \mathcal{L}_{\text{direction}}\left(\mathbf{P}_{t}^{\text{edit}}, c^{\text{target}} ; \mathbf{P}_{t}^{\text{source}}, c^{\text{source}}\right) \\
    + \lambda_{\text{reg}} |\mathbf{P}_{t}^{\text{edit}} - \mathbf{P}_{t}^{\text{source}}|.
\end{multline}

\subsubsection{Adversarial Guidance}
We also introduce an adversarial guidance to the semantic function $\mathbf{F}$ to learn a semantic attribute but is adversarial to the target classifier $\boldsymbol{f}$.
We focus on targeted unrestricted adversarial attacks, where the attacker tries to generate an image $\boldsymbol{x}$ that looks like the ground-truth label $y_{\text{source}}$ but is misclassified as 
the target label $y_{\text{target}}$ by $\boldsymbol{f}$.
Different from previous works~\cite{17,18} feeding the intermediate latent noise $\boldsymbol{x}_t$ to the target model to obtain the adversarial guidance at each timestep, we adopt the modified $\mathbf{P}_{t}(\tilde{\boldsymbol{\epsilon}}_t^\theta)$ for better approximation of the final output $\boldsymbol{x}_0$:
\begin{equation}
    \mathcal{L}_{\text{adv}} = \mathcal{L}_{\text{CE}}\left(\boldsymbol{f}(\mathbf{P}_{t}(\tilde{\boldsymbol{\epsilon}}_t^\theta(\boldsymbol{x}_t))), y_{\text{target}}\right),
\end{equation}
where $\mathcal{L}_{\text{CE}}$ is the cross-entropy loss of the target model.

Hence, the overall loss of the semantic function $\mathbf{F}$ is:
\begin{equation}
    \mathcal{L} = \mathcal{L}_{\text{CLIP}} + \lambda_{\text{adv}} \mathcal{L}_{\text{adv}}.
\end{equation}
For each attribute $\boldsymbol{a}_i$, we train a specific semantic function $\mathbf{F}_{i}$ in the same way, as depicted in Fig.~\ref{fig2}. The training process is detailed in Algorithm~\ref{alg1}.

\begin{algorithm}[tbp]
    \caption{Training Semantic Function $F(h_{t}, t)$}
    \label{alg1}
    \KwIn{$\epsilon_{\theta}$(pretrained model), $\{x_0^{(i)}\}_{i=1}^N$(images to precompute), $F$(Semantic function of an attribute), $c^{src}$(source text), $c^{tar}$(target text), $S_{inv}$(\# of inversion steps), $K$(\# of training epochs), $f$(target model), $y_{tar}$(target label)}
    \KwOut{$F$ (trained semantic function)}
    \BlankLine
    \tcp{Step 1: Precompute latents}
    Define $\{\tau_s\}_{s=1}^{S_{inv}}$ s.t $\tau_1 = 0, \tau_{S_{inv}} = T$\;
    \For{$i = 1, 2, \ldots, N$}{
        \For{$s = 1, 2, \ldots, S_{inv} - 1$}{
            $\epsilon \leftarrow \epsilon_{\theta}(x_{\tau_s}^{(i)}, \tau_s)$\;
            $x_{\tau_{s+1}}^{(i)} = \sqrt{\alpha_{\tau_s}} x_{\tau_s}^{(i)} + \sqrt{1 - \alpha_{\tau_s}} \epsilon$\;
        }
        Save the latent $x_{\tau_{S_{inv}}}^{(i)}$\;
    }
    \BlankLine
    \tcp{Step 2: Update $F$}
    \For{epoch $= 1, 2, \ldots, K$}{
        \For{$i = 1, 2, \ldots, N$}{
            Clone the latent $\tilde{x}_{\tau_{S_{inv}}}^{(i)} = x_{\tau_{S_{inv}}}^{(i)}$\;
            \For{$s = S_{inv}, S_{inv} - 1, \ldots, 1$}{
                Extract feature map $h_{\tau_s}$ from $\epsilon_{\theta}(\tilde{x}_{\tau_s}^{(i)})$\;
                $\Delta h_{\tau_s} = F(h_{\tau_s}, \tau_s)$\;
                $P = \frac{\tilde{x}_{\tau_s}^{(i)} - \sqrt{1 - \alpha_{\tau_s}} \epsilon_{\theta}(\tilde{x}_{\tau_s}^{(i)}) \Delta h_{\tau_s}}{\sqrt{\alpha_{\tau_s}}}; P_{src} = \frac{x_{\tau_s}^{(i)} - \sqrt{1 - \alpha_{\tau_s}} \epsilon_{\theta}(x_{\tau_s}^{(i)})}{\sqrt{\alpha_{\tau_s}}}$\;
                $\tilde{x}_{\tau_{s-1}}^{(i)} = \sqrt{\alpha_{\tau_{s-1}}} P + \sqrt{1 - \alpha_{\tau_{s-1}}} \epsilon_{\theta}(\tilde{x}_{\tau_s}^{(i)})$\;
                $x_{\tau_{s-1}}^{(i)} = \sqrt{\alpha_{\tau_{s-1}}} P_{src} + \sqrt{1 - \alpha_{\tau_{s-1}}} \epsilon_{\theta}(x_{\tau_s}^{(i)})$\;
            }
            $L \leftarrow L_{direction}(P, c^{tar}, P_{src}, c^{src}) + \lambda_{recon} \|P - P_{src}\| + \lambda_{adv} L_{CE}(f(P), y_{tar}) $\;
            Take a gradient step on $L$ and update $F$\;
        }
    }
\end{algorithm}

\subsection{Multi-attributes Optimization Approach}
Given a trained semantic function $\mathbf{F}$, we can generate $\Delta \boldsymbol{h}_{t} = \boldsymbol{w}_{t} \mathbf{F}(\boldsymbol{h}_{t}, t)$ to derive the UAE $\boldsymbol{x}_0$.
However, a single attribute usually needs a large weight $\boldsymbol{w}_{t}$ to successfully deceive target models, which results in unnatural UAEs with excessive semantic shifts.
To address this issue, we further propose a multi-attributes optimization approach to optimize the weights of multiple adversarial semantic attributes with a weight penalty term, so that slight changes of multiple attributes are sufficient to cause misclassification while keeping UAEs natural and close to the sampled images.

With a set of attributes $\{\boldsymbol{a}_i\}_{i=1}^M$ and their corresponding semantic functions $\{\mathbf{F}_{i}\}_{i=1}^M$, we optimize the weights $\boldsymbol{w}_{it}$ of different attributes to minimize the following objective:
\begin{equation}
    \begin{aligned}
        \min_{\boldsymbol{w}_{it}} \ & \mathcal{L}_{\text{CE}}(\boldsymbol{f}(\mathbf{P}_{t}^{edit}), \ y_{\text{target}}) + \lambda_{1} \mathcal{L}_{\text{CE}}(\boldsymbol{g}(\mathbf{P}_{t}^{edit}), \ y_{\text{source}}) \\
        & + \lambda_{2} \sum |\boldsymbol{w}_{it}|,
    \end{aligned}
\end{equation}
where $\mathbf{P}_{t}^{edit} = \mathbf{P}_t\left(\boldsymbol{\epsilon}_t^\theta\left(\boldsymbol{x}_t \mid \sum \boldsymbol{w}_{it} \mathbf{F}_{i}(\boldsymbol{h}_{it}, t)\right)\right)$, the first cross-entropy loss encourages $\boldsymbol{f}$ to predict $y_{\text{target}}$, the second cross-entropy loss encourages another auxiliary classifier $\boldsymbol{g}$ to make correct prediction $y_{\text{source}}$, and the third term penalizes $\boldsymbol{w}_{it}$ to prevent excessive semantic shifts. We hypothesize that $\boldsymbol{g}$ is relatively uncorrelated with $\boldsymbol{f}$, that can possibly promote the generated UAEs to reside in class $y_{\text{source}}$ while crossing the decision boundary of $\boldsymbol{f}$.

Considering diffusion models generate high-level context in the early stage and refine details in the later stage~\cite{25}, we only modify the generative process in the early stage to achieve adversarial semantic changes with the editing interval [$T, t_{edit}$], and inject stochastic noise in the later stage to boost image quality with the interval [$t_{boost}, 0$].
The modified generative process of DDIM $p_\theta^{(t)}\left(\boldsymbol{x}_{t-1} \mid \boldsymbol{x}_t\right)$ becomes:

\begin{equation}
    \begin{cases}
        \sqrt{\alpha_{t-1}} \mathbf{P}_t\left(\boldsymbol{\epsilon}_t^\theta\left(\boldsymbol{x}_t \mid \Delta \boldsymbol{h}_{t}\right)\right)+\mathbf{D}_t & \text { if } T \geq t \geq t_{\text {edit }} \\ 
        \sqrt{\alpha_{t-1}} \mathbf{P}_t\left(\boldsymbol{\epsilon}_t^\theta\left(\boldsymbol{x}_t\right)\right)+\mathbf{D}_t & \text { if } t_{\text {edit }}>t \geq t_{\text {boost }} \\ 
        \mathcal{N}\left(\sqrt{\alpha_{t-1}} \mathbf{P}_t\left(\boldsymbol{\epsilon}_t^\theta\left(\boldsymbol{x}_t\right)\right)+\mathbf{D}_t, \sigma_t^2 \boldsymbol{I}\right) & \text { if } t_{\text {boost }}>t
    \end{cases}
\end{equation}
 where $\eta=1$ only in the interval [$t_{\text {boost}}, 0$].
 We follow Kwon et al.~\cite{24} to find $t_{\text {edit}}$ with $\text {LPIPS}(\boldsymbol{x}, \mathbf{P}_{t_{\text {edit}}})=0.33 - \delta$ which represents the shortest editing interval from $T$ to $t$ when $\mathbf{P}_t$ approaches the original image $\boldsymbol{x}$.
 $\delta$ is adaptive to the cosine distance of CLIP text embeddings for different attributes, namely $\delta = 0.33 d(E_T\left(c^{\text{source}}\right), E_T\left(c^{\text{target}}\right))$.
 Similarly, we find $t_{\text {boost}}$ with $\text {LPIPS}(\boldsymbol{x}, \boldsymbol{x_t}) = 1.2$ which achieves quality boosting with minimal content change.
 
 Starting from a sampled noise $\boldsymbol{x}_T$, we progressively denoise it through the above three phases to obtain the final output $\boldsymbol{x}_0$ as the potential adversarial example.
 The SemDiff unrestricted adversarial attack is detailed in Algorithm~\ref{alg2}.

\begin{algorithm}[tp]
    \caption{The algorithm of SemDiff}
    \label{alg2}
    \KwIn{$\epsilon_{\theta}$(frozen pretrained model), $f$(target classifier), $g$(auxiliary classifier), $\{F_{i}\}_{i=1}^M$($M$ Semantic functions for $M$ attributes), $S_{gen}$(\# of inference steps), $K$(\# of iterations), $y_{tar}$(target label), $y_{src}$(source label), $t_{edit}$(timestep for semantic editing), $t_{boost}$(timestep for quality boosting)}
    \KwOut{$x_0$ (adversarial example)}
    \BlankLine
    Define \(\{\tau_s\}_{s=1}^{S_{gen}}\) s.t \(\tau_1 = 0, \tau_{S_{edit}} = t_{edit}, \tau_{S_{noise}} = t_{\text{boost}}\) and \(\tau_{S_{gen}} = T\)\;
    \(x_{\tau_{S_{gen}}} \sim \mathcal{N}(0, 1)\); \(w_{i} \sim \mathcal{N}(0, 1)\)\;
    \For{$iter = 1, 2, \ldots, K$}{
        \For{\(s = S_{gen}, S_{gen} - 1, \dots, 2\)}{
            \tcp{Phase 1: Semantic editing}
            \If{\(s \geq S_{edit}\)}{
                Extract feature map \(h_{\tau_s}\) from \(\epsilon_\theta(x_{\tau_s})\)\;
                \(\Delta h_{\tau_s} = \sum_{i=1}^M w_{i\tau_s} F_{i} (h_{\tau_s}, \tau_s)\)\;
                \(\tilde{\epsilon} = \epsilon_\theta(x_{\tau_s} | \Delta h_{\tau_s})\)\;
                \(\epsilon = \epsilon_\theta(x_{\tau_s})\)\;
                \(\sigma_{\tau_s} = 0\)\;
                \(\tilde{P}_{\tau_s} = \frac{x_{\tau_s}-\sqrt{1-\alpha_{\tau_s}\tilde{\epsilon}}}{\sqrt{\alpha_{\tau_s}}} \)\;
                \(L \leftarrow L_{\text{CE}}(f(\tilde{P}_{\tau_s}), \ y_{\text{tar}}) + \lambda_{1} L_{\text{CE}}(g(\tilde{P}_{\tau_s}), \ y_{\text{src}}) + \lambda_{2} \sum |\boldsymbol{w}_{i\tau_s}|\)\;
                Take a gradient step on $L$ and update $w_{i\tau_s}$\;
            }
            \tcp{Phase 2: Denoising}
            \ElseIf{\(s \geq S_{noise}\)}{
                \(\tilde{\epsilon} = \epsilon = \epsilon_\theta(x_{\tau_s})\)\;
                \(\sigma_{\tau_s} = 0\)
            }
            \tcp{Phase 3: Quality boosting}
            \Else{
                \(\tilde{\epsilon} = \epsilon = \epsilon_\theta(x_{\tau_s})\)\;
                \(\sigma_{\tau_s} = \sqrt{\frac{1-\alpha_{\tau_s-1}}{1-\alpha_{\tau_s}}} \sqrt{1 - \frac{\alpha_{\tau_s}}{\alpha_{\tau_s-1}}}\)\;
            }
            \(z \sim \mathcal{N}(0, 1)\)\;
            \(x_{\tau_s-1} = \sqrt{\alpha_{\tau_s-1}} (\frac{x_{\tau_s}-\sqrt{1-\alpha_{\tau_s}\tilde{\epsilon}}}{\sqrt{\alpha_{\tau_s}}}) + \sqrt{1-\alpha_{\tau_s-1} -\sigma_{\tau_s}^2} \epsilon + \sigma_{\tau_s} z\)\;
        }
        \If{\(f(x_0) = y_{\text{tar}}\)}{
            \Return{\(x_0\)}
        }
    }
\end{algorithm}

\section{Experiments}
\noindent In this section, we conduct extensive experiments to demonstrate the effectiveness of the proposed SemDiff.
We first introduce the experiment settings. 
Then we evaluate SemDiff in unrestricted adversarial attack on four tasks and three high-resolution datasets.
Following, we further investigate the efficacy and imperceptibility of the generated UAEs by testing SemDiff against different defense methods.
Finally, we provide the ablation study of SemDiff and report its hyperparameter analysis.

\subsection{Experiment Settings}
\subsubsection{Datasets and Target Models}
We evaluate unrestricted adversarial attacks on four tasks on three high-resolution datasets, including gender classification on CelebA-HQ~\cite{37}, animal classification on AFHQ~\cite{38}, church classification and any-class classification on ImageNet~\cite{1}.
For the targeted classifiers to attack, we use the pretrained ResNet50~\cite{39} and ViT~\cite{40} for ImageNet dataset, and retrain them for the tasks on CelebA-HQ and AFHQ.
We also adopt VGG16~\cite{41} as the auxiliary classifier for our SemDiff attack and the baseline UGAN~\cite{12} attack.
The performance of the classifiers are reported in Tab.~\ref{tab:classifier_performance}.
\begin{table}[tbp]
	\caption{Accuracy of the classifiers} \label{tab:classifier_performance}
	\centering
	\begin{tabular}{c|ccc}
        \hline
        {Classifier} & {CelebA-HQ} & {AFHQ} & {ImageNet}\\
        \hline
        {ResNet50} & 98.4\% & 99.2\% & 76.1\%\\
        {ViT} & 97.4\% & 99.7\% & 81.1\%\\
        \hline
        {VGG16} & 98.6\% & 99.8\% & 71.6\%\\
        \hline
	\end{tabular}
\end{table}

\subsubsection{Baselines}
We compare our proposed SemDiff with the state-of-the-art unrestricted adversarial attacks, including GAN-based UGAN~\cite{12}, diffusion-based AdvDiff~\cite{17} and AdvDiffuser~\cite{18}.
For a fair comparison, we adopt the same pretrained diffusion models for diffusion-based attacks, namely SDEdit~\cite{22} on CelebA-HQ and AFHQ, ADM~\cite{20} on ImageNet.
For GAN-based UGAN, we use StyleGAN2~\cite{42} on CelebA-HQ and AFHQ, BigGAN~\cite{43} on ImageNet.

\subsubsection{Evaluation Metrics}
We use the Attack Success Rate (ASR) to evaluate the attack performance.
In terms of the imperceptibility and naturalness of the generated UAEs, we utilize the Fr\'echet Inception Distance (FID)~\cite{44}, Kernel Inception Distance (KID)~\cite{45} and BRISQUE~\cite{46}.
Among them, FID and KID measure the similarity between generated UAEs and real-world images, while BRISQUE is a no-reference image quality assessment metric that can evaluate image naturalness similar to human perception.

\subsubsection{Implementation Details}
We perform targeted attacks in all experiments.
For each attack, we generate 1000 UAEs with the same sampled noise images for diffusion models.
Since GANs use different input noise vector, we reconstruct the sampled noise images of diffusion models, and then project them to the input latent space of GANs as the input vectors for a fair comparison.
For our method, we train $\boldsymbol{F}$ for 1 epoch using 1,000 samples with $S=40$ on CelebA-HQ and AFHQ, $S=80$ on ImageNet.
We set $\lambda_{adv}=1$, and use $\lambda_{reg}= \text{CLIP simlarity} * 3$ when reducing L1 loss for different attributes.
For multi-attributes optimization, we use $\lambda_{1}=1$ and $\lambda_{2}=10$ on CelebA-HQ and AFHQ, $\lambda_{1}=2$, $\lambda_{2}=15$ on ImageNet.
We set $K=40$ on each attack for a fair comparison.
More details about the hyperparameters $\lambda_{adv}$, $\lambda_{reg}$, $t_{edit}$, $t_{boost}$ and $\text{CLIP simlarity}$ can be found in Supplementary Material B.
We devise a set of attributes with source/target text pairs for each task in Tab.~\ref{tab:attributes}.
Some attributes are neutral such as \textit{smiling} and \textit{young}, while some attributes correspond to the target classes, such as \textit{strong jawline} and \textit{busy eyebrows} for male class in gender classification task.
The code is available at: https://github.com/Daizy97/SemDiff.
\begin{table}[tbp]
	\caption{Attributes of different tasks} \label{tab:attributes}
	\centering
    \resizebox{\columnwidth}{!}{ 
	\begin{tabular}{c|ccc}
        \hline
        {Task} & {attribute} & {source text} & {target text}\\
        \hline
        \multirow[c]{8}{*}{Gender} & {smiling} & {face} & {smiling face}\\
        & {young} & {person} & {young person}\\
        & {tanned} & {face} & {tanned face}\\
        & {short hair} & {person} & {person with short hair}\\
        & {strong jawline} & {person} & {person with strong jawline}\\
        & {busy eyebrows} & {person} & {person with busy eyebrows}\\
        & {big round eyes}  & {person} & {person with big round eyes}\\
        & {soft facial lines} & {person} & {person with soft facial lines}\\
        \hline
        \multirow[c]{4}{*}{Animal} & {dog happy} & {dog} & {happy dog}\\
        & {dog cat fur} & {dog} & {dog with fur of cat}\\
        & {dog cat nose} & {dog} & {dog with short snout of cat}\\
        & {dog cat ears} & {dog} & {dog with pointed ears of cat}\\
        \hline
        \multirow[c]{3}{*}{Church} & {church night} & {church} & {church at night}\\
        & {church old} & {church} & {old church}\\
        & {church wooden} & {church} & {wooden church}\\
        \hline
        \multirow[c]{6}{*}{Any-class} & {sunny} & {photo} & {photo of a sunny day}\\
        & {rainy} & {photo} & {photo of a rainy day}\\
        & {foggy} & {photo} & {photo of a foggy day}\\
        & {motion blur} & {photo} & {photo with motion blur}\\
        & {defocus blur} & {photo} & {photo with defocus blur}\\
        & {frosted glass blur} & {photo} & {photo with frosted glass blur}\\
        \hline
	\end{tabular}
    }
\end{table}

\subsection{Unrestricted Adversarial Attack}
\label{subsec:comparison_experiment}
\subsubsection{Gender Classification on CelebA-HQ}
In this task, the attack aims to generate UAEs that look like females but are misclassified as males.
Our proposed SemDiff uses $\{$\textit{smiling}, \textit{young}, \textit{tanned}, \textit{strong jawline}, \textit{busy eyebrows}$\}$ as the attribute set.
The detailed results are presented in Tab.~\ref{tab:comparison_results}.
It is observed that all three diffusion-based attack can achieve 100\% ASR, except for UGAN-StyleGAN2, which demonstrates the effectiveness of diffusion models in generating UAEs.
Moreover, SemDiff shows the superior performance in terms of BRISQUE, FID and KID, indicating the generated UAEs of SemDiff are more natural and realistic.

\begin{figure}[tbp]
    \centering
    \includegraphics[width=\columnwidth]{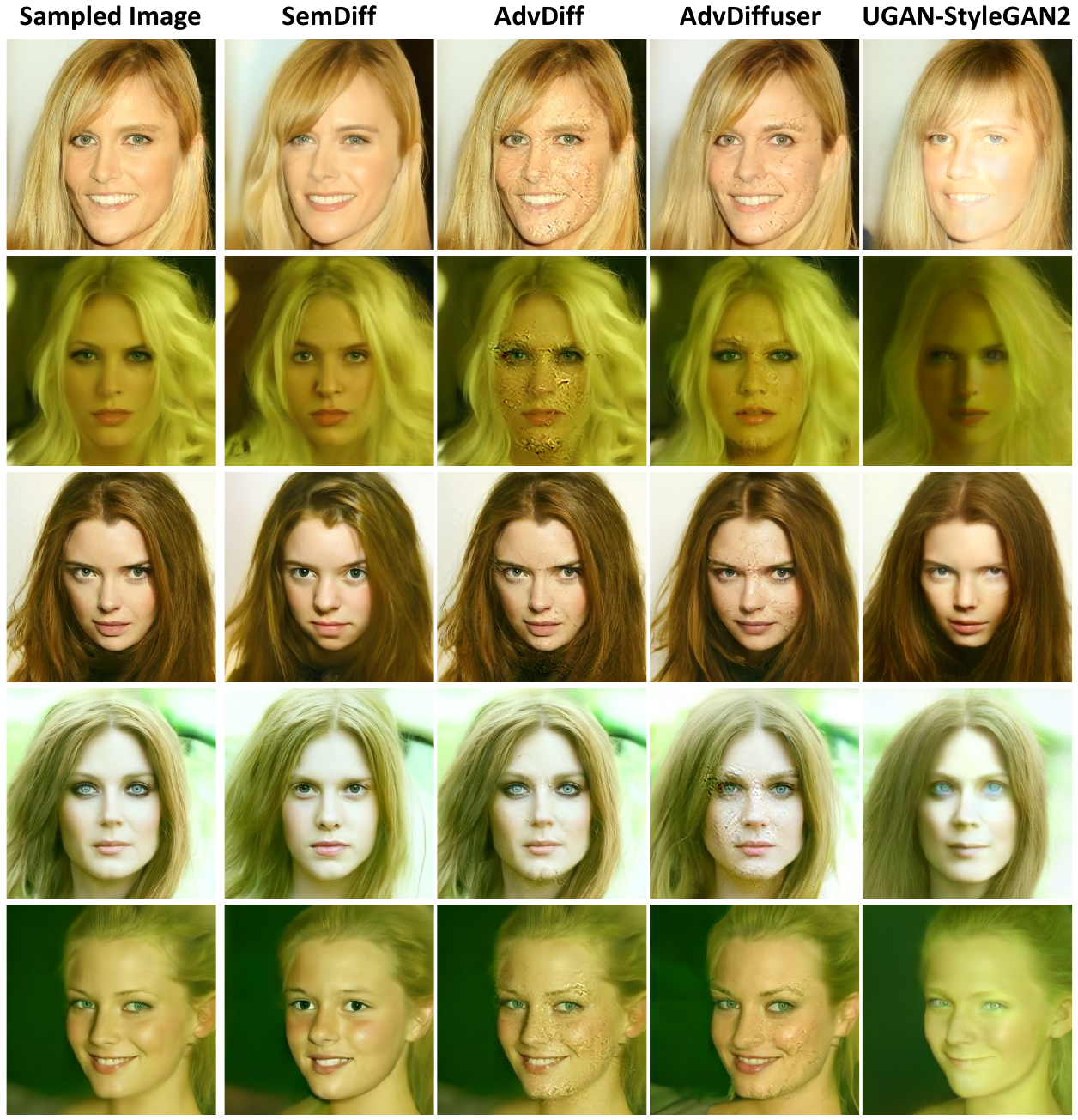}
    \caption{Visual examples of the UAEs generated by different attacks for gender classification task on CelebA-HQ. The first column shows the sampled images reconstructed with the randomly sampled noise images by diffusion models.}
    \label{fig3}
\end{figure}

We also provide some visual examples in Fig.~\ref{fig3}, where the first column shows the sampled images reconstructed with the randomly sampled noise images by diffusion models.
Notice that the sampled images are just the output images if we do not modify the generative models for attack.
We can find that the UAEs generated by AdvDiff and AdvDiffuser contain perceptible perturbations and basically do not change any semantic contents compared with the sampled images.
This illustrates that current diffusion-based attacks that perturb in the intermediate latent noise space perform like traditional perturbation-based attacks with superficial perturbations, resulting in inadequate imperceptibility and naturalness.
Instead, our SemDiff generate natural UAEs with meaningful semantic changes consistent with the attribute set, such as becoming younger for the females in the images.
Such slight and meaningful semantic changes accord with our cognition thus more imperceptible to human eyes.
Although UGAN-StyleGAN2 also reveals some semantic changes in their UAEs, their semantic directions are unknown and cannot guarantee naturalness because UGAN attack just optimizes the input latent vector in GANs.
In comparison, SemDiff optimizes the weights of different semantic attributes, making the generated UAEs more controllable and realistic.

\begin{table*}[tbp]
	\caption{Comparison results of the proposed SemDiff attack with baselines in unrestricted adversarial attack} \label{tab:comparison_results}
	\centering
	\begin{tabular}{c|c|cccc|cccc}
        \hline
        \multirow[c]{2}{*}{Task} & \multirow[c]{2}{*}{Method} & \multicolumn{4}{c|}{ResNet50} &  \multicolumn{4}{c}{ViT}\\
        & & {ASR↑} & {BRISQUE↓} & {FID↓} & {KID↓} & {ASR↑} & {BRISQUE↓} & {FID↓} & {KID↓}\\
        \hline
        \multirow[c]{4}{*}{Gender classification on CelebA-HQ} & {AdvDiff} & \textbf{100\%} & 27.28 & 28.81 & 0.009 & \textbf{100\%} & 33.64 & 22.70 & 0.008\\
        & {AdvDiffuser} & \textbf{100\%} & 27.27 & 28.33 & 0.009& \textbf{100\%} & 33.64 & 22.21 & 0.008\\
        & {UGAN-StyleGAN2} & 93.2\% & 31.79 & 38.75 & 0.014 & 95.0\% & \textbf{30.79} & 39.61 & 0.015\\
        & {\textbf{SemDiff (ours)} } & \textbf{100\%} & \textbf{26.59} & \textbf{27.60} & \textbf{0.007} & \textbf{100\%} & 31.37 & \textbf{19.77} & \textbf{0.007}\\
        \hline
        \multirow[c]{4}{*}{Animal classification on AFHQ} & {AdvDiff} & \textbf{96.9\%} & 24.93 & 30.46 & 0.015 & 93.7\% & 27.68 &39.59 &0.018\\
        & {AdvDiffuser} & 96.3\% & 25.22 & 30.37 & 0.015 & 93.9\% & 27.88 & 39.38 & 0.018\\
        & {UGAN-StyleGAN2} & 91.7\% & 26.03 & 38.40 & 0.024 & 88.9\% & 28.00 & 47.08 & 0.036\\
        & {\textbf{SemDiff (ours)} } & 95.4\% & \textbf{23.98} & \textbf{26.40} & \textbf{0.010} & \textbf{94.6\%} & \textbf{24.59} & \textbf{36.36} & \textbf{0.012}\\
        \hline
        \multirow[c]{4}{*}{Church classification on ImageNet} & {AdvDiff} & \textbf{98.9\%} & 28.09 & 31.70 & 0.015 & 97.6\% & 31.21 & \textbf{36.31} & \textbf{0.016}\\
        & {AdvDiffuser} & 98.2\% & 28.23 & 31.76 & 0.015 & 96.9\% & 31.03 & 36.38 & \textbf{0.016}\\
        & {UGAN-BigGAN} & 94.3\% & 29.01 & 38.40 & 0.021 & 92.7\% & 31.23 & 45.08 & 0.028\\
        & {\textbf{SemDiff (ours)} } & 98.4\% & \textbf{26.14} & \textbf{28.55} & \textbf{0.014} & \textbf{98.2\%} & \textbf{26.02} & 38.31 & 0.017\\
        \hline
        \multirow[c]{4}{*}{Any-class classification on ImageNet} & {AdvDiff} & \textbf{99.1\%} & 27.47 & 33.57 & 0.020 & \textbf{98.3\%} & 30.21 & 35.31 & 0.019\\
        & {AdvDiffuser} & 97.9\% & 27.03 & 34.86 & 0.020 & 97.2\% & 30.31 & 34.23 & 0.019\\
        & {UGAN-BigGAN} & 94.7\% & 33.02 & 46.97 & 0.037 & 83.4\% & 33.55 & 41.30 & 0.032\\
        & {\textbf{SemDiff (ours)} } & 98.5\% & \textbf{25.04} & \textbf{29.62} & \textbf{0.017} & 97.4\% & \textbf{27.74} & \textbf{28.05} & \textbf{0.017}\\
        \hline
	\end{tabular}
\end{table*}

\subsubsection{Animal Classification on AFHQ}
This task have three classes, including dog, cat and wildlife.
The attack aims to generate UAEs that look like dogs but are misclassified as cats.
We use $\{$\textit{dog happy}, \textit{dog cat fur}, \textit{dog cat nose}, \textit{dog cat ears}$\}$ as the attribute set for SemDiff.
From Tab.~\ref{tab:comparison_results}, we can observe that our SemDiff outperforms all the baselines in terms of BRISQUE, FID and KID, confirming the imperceptibility and naturalness of SemDiff.
AdvDiff obtains slightly higher ASR than other attacks for ResNet50 as the target classifier.
However, when faced with stronger ViT, the ASR of all attacks decreases, while SemDiff achieves the highest ASR, which demonstrates the attack effectiveness of SemDiff.

\begin{figure}[tbp]
    \centering
    \includegraphics[width=\columnwidth]{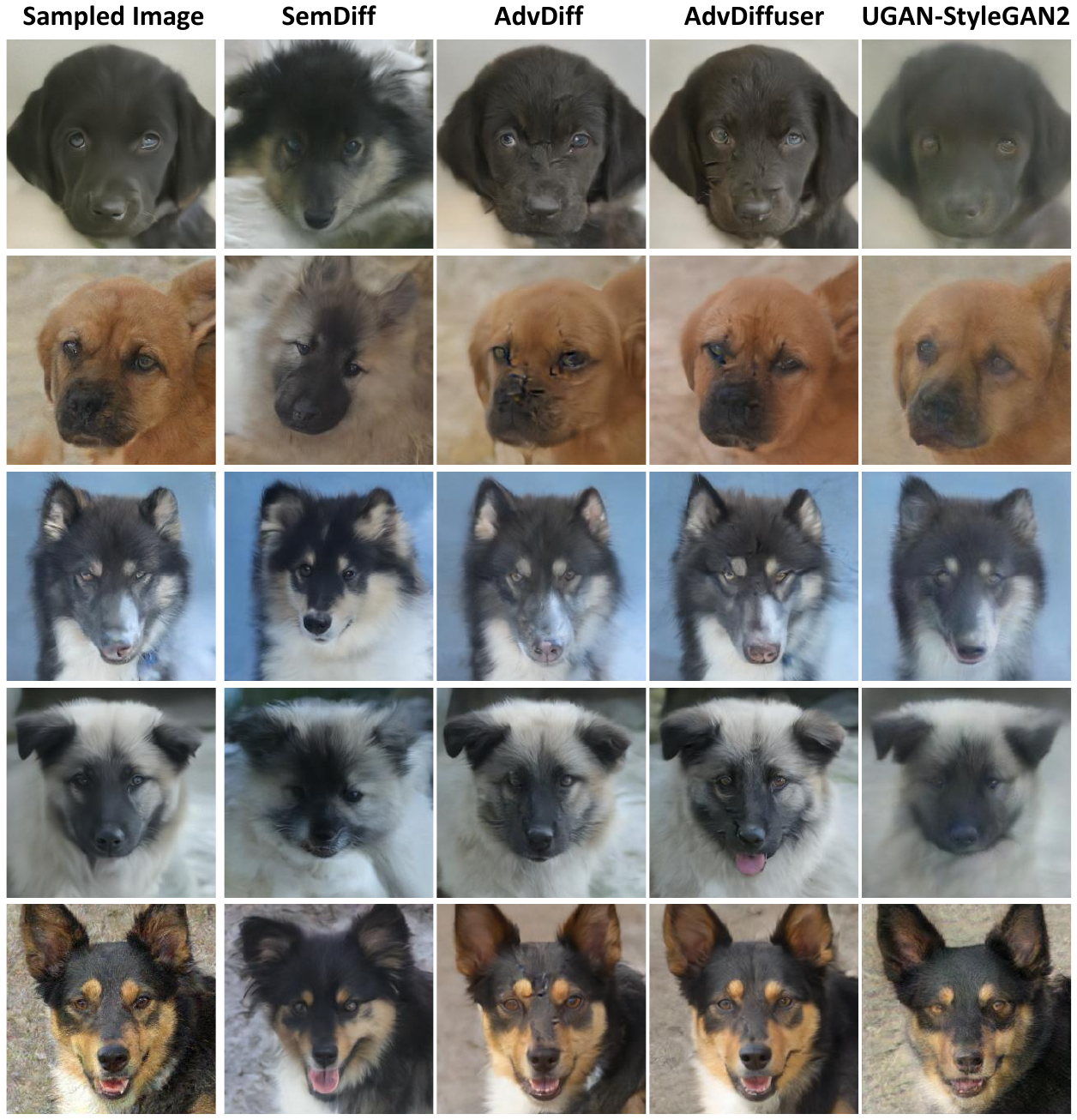}
    \caption{Visual examples of the UAEs generated by different attacks for animal classification task on AFHQ. The first column shows the sampled images reconstructed with the randomly sampled noise images by diffusion models.}
    \label{fig4}
\end{figure}

For the visual examples in Fig.~\ref{fig4}, we notice the similar phenomenon in CelebA-HQ that the UAEs generated by AdvDiff and AdvDiffuser show limited naturalness with visible perturbations, while SemDiff and UGAN-StyleGAN2 modify the semantics of UAEs to cause misclassification.
Furthermore, SemDiff exhibits more natural and meaningful semantic changes compared with UGAN-StyleGAN2, leading to UAEs of higher quality.
For example, some dogs have denser hair and their ears become shorter and more pointed, resembling those of cats.
These semantic changes also accord with \textit{dog cat fur} and \textit{dog cat ears} attributes.

\subsubsection{Church classification on ImageNet}
ImageNet contains 1000 classes, and this task aims to generate UAEs that look like churches but are misclassified as other classes.
We use $\{$\textit{church night}, \textit{church old}, \textit{church wooden}$\}$ as the attribute set for SemDiff.
According to the results in Tab.~\ref{tab:comparison_results}, SemDiff obtains comparable performance with AdvDiff in terms of ASR and better performance in most of the naturalness metrics, which demonstrates the better naturalness of SemDiff once again.
We also observe that there is a relatively large gap between UGAN-BigGAN and diffusion-based attacks in terms of ASR, FID and KID, which suggests that the performance of unrestricted adversarial attacks is heavily affected by the generation quality of generative models.
More visual examples can be found in Supplementary Material C.

\subsubsection{Any-class Classification on ImageNet}
In previous tasks, the attributes set is designed specific to the source class, which may limit the generalization of the proposed SemDiff to other classes.
In this task, we try to create some universal attributes that can be applied to any class for modification.
Inspired by the common visual corruptions used in ImageNet-C dataset~\cite{47}, we design a set of attributes related to some common weather or blurs that usually exist in a photo: $\{$\textit{sunny}, \textit{foggy}, \textit{motion blur}, \textit{defocus blur}, \textit{frosted glass blur}$\}$.
Therefore, any class object can be modified with these universal attributes to achieve attack.
The results are reported in Tab.~\ref{tab:comparison_results}.
It is found that SemDiff shows comparable performance with other diffusion-based attacks, and outperforms UGAN-BigGAN with a large margin in terms of ASR.
In addition, SemDiff still dominates other attacks in all three imperceptibility and naturalness metrics, which confirms the superiority of our attack.
We also provide visual examples in Supplementary Material C.

\begin{table}[tbp]
	\caption{Comparison results of the proposed SemDiff attack with baselines against four defenses} \label{tab:defense}
	\centering
    \resizebox{\columnwidth}{!}{ 
	\begin{tabular}{c|c|ccccc}
        \hline
        {Task} & {Method} & {None} & {JPEG} &  {FS} & {SRNet} & {AT}\\
        \hline
        \multirow[c]{4}{*}{ResNet50} & {AdvDiff} & 99.1\% & 50.1\% & 60.5\% & 55.1\% & 58.7\%\\
        & {AdvDiffuser} & 97.9\% & 52.4\% & 58.3\% & 54.8\% & 59.4\%\\
        & {UGAN} & 94.7\% & 65.8\% & 70.6\% & 52.3\% & 55.1\%\\
        & {\textbf{SemDiff} } & 98.8\% & \textbf{71.3\%} & \textbf{81.4\%} & \textbf{56.2\%} & \textbf{64.9\%}\\
        \hline
        \multirow[c]{4}{*}{ViT} & {AdvDiff} & 98.3\% & 56.2\% & 55.1\% & 53.7\% & 56.4\%\\
        & {AdvDiffuser} & 97.2\% & 58.7\% & 52.4\% & \textbf{58.4\%} & 60.2\%\\
        & {UGAN} & 83.4\% & 66.1\% & 68.4\% & 49.6\% & 58.3\%\\
        & {\textbf{SemDiff} } & 97.4\% & \textbf{78.3\%} & \textbf{73.9\%} & 54.7\% & \textbf{63.8\%}\\
        \hline
	\end{tabular}
    }
\end{table}
\subsection{Adversarial Defenses}
To further validate the efficacy of the proposed SemDiff attack, we investigate whether our SemDiff can evade four different defense methods, including JPEG compression~\cite{48}, feature squeezing~\cite{49}, SRNet~\cite{50} and adversarial training model~\cite{51}.
JPEG compression is a preprocessing-based defense that aims to eliminate adversarial perturbations by compressing input images~\cite{48}.
Feature squeezing (FS)~\cite{49} and SRNet~\cite{50} are two detection-based defenses that try to detect adversarial examples before feeding them into the classifier, where FR detects adversarial examples based on the uncertainty of model output before and after squeezing, SRNet is a CNN-based binary classifier to distinguish adversarial examples from clean ones.
Adversarial training (AT)~\cite{52} is a robustness-based defense that trains the classifier with adversarial examples to improve the robustness of the model against adversarial attacks.
We set the quality factor as 75 for JPEG compression, ($5-bit, 2\times2, 11-3-4$) for FS, finetune SRNet with the UAEs generated by all attacks, and use an adversarially training model IncRes-v2-adv-ens~\cite{51}.

The evaluation results are presented in Tab.~\ref{tab:defense}, where the None column denotes the ASR without any defense while the other columns represent the ASR under different defense methods.
It is found that the proposed SemDiff is more robust than the baseline attacks against most of the defenses.
Specifically, the SemDiff attack achieves the highest ASR under JPEG compression, feature squeezing, and adversarial training with a large margin.
Among them, JPEG compression and feature squeezing both preprocess input images with compression, smoothing or color bit squeezing, which can remove high-frequency noise and perturbations in images.
The adversarial examples used for adversarial training are generated by perturbation-based attacks, such as FGSM~\cite{6}.
As discussed before, the generated UAEs of AdvDiff and AdvDiffuser perform like traditional perturbation-based attacks with superficial perturbations, thus are more sensitive to these defenses in comparison with SemDiff and UGAN.
For the binary classifier detector SRNet which is trained with the UAEs generated by all test attacks, the ASR degrades most significantly, but SemDiff can still obtain a competitive performance.
In summary, the proposed SemDiff attack is more robust against various defenses, which further confirms its attack effectiveness and imperceptibility.

\subsection{Ablation Study}
In this subsection, we carry out the ablation study on the proposed SemDiff method, including the adversarial semantic attribute and the multi-attributes optimization approach.
To verify the effectiveness of the adversarial semantic attribute, we additionally train the semantic function $\boldsymbol{F}$ without adversarial guidance to obtain clean semantic attributes.
As for the multi-attributes optimization approach, we implement a single-attribute optimization using line search to gradually increase the attribute weight until misclassification.

We adopt the SemDiff on CelebA-HQ with the attribute set $\{$\textit{smiling}, \textit{young}, \textit{tanned}, \textit{strong jawline}, \textit{busy eyebrows}$\}$ for the ablation study.
We use ``adv\_attr'' and ``attr'' to distinguish whether the method employs adversarial semantic attributes or clean semantic attributes, and use ``multi'' and ``single'' to distinguish if the method employs multi-attributes optimization or single-attribute optimization.
The results are provided in Tab.~\ref{tab:ablation}.
We report an extra metric ``Weight'' to measure the mean absolute weight of all attributes when the attack succeeds.
The weight is expected to be close to 0 so that the generated UAEs resemble the sampled images with slight semantic changes but can successfully mislead the target classifier.

\subsubsection{Effectiveness of Adversarial Semantic Attribute}
From Tab.~\ref{tab:ablation}, we can find that all ``adv\_attr'' methods gain higher ASR than the corresponding ``attr'' methods, which demonstrates the adversarial semantic attributes indeed enhance the attack capability compared with clean semantic attributes.
Moreover, the weight of attributes of all ``adv\_attr'' methods is also much smaller than that of the corresponding ``attr'' methods, indicating that the SemDiff equipped adversarial semantic attributes can achieve attack success with slighter semantic changes and higher efficiency.
On the other hand, two types of methods have similar BRISQUE performance, but ``adv\_attr'' methods achieve better FID and KID.
This suggests that although clean semantic attributes can preserve the naturalness of UAEs to some extent, they generally underperform compared to adversarial semantic attributes, as the UAEs that are difficult to attack successfully will exhibit more significant and even distorted semantic changes.
Therefore, the results validate the efficacy of adversarial semantic attributes in the SemDiff attack.

\subsubsection{Effectiveness of Multi-Attributes Optimization Approach}
We can observe that in Tab.~\ref{tab:ablation}, ``adv\_attr \& multi'' method performs better than any ``single'' method, especially in terms of the attribute weight.
Due to the limited optimization space of the single-attribute optimization method, it is hard to achieve the same level of attack efficiency as the multi-attributes optimization method.
Meanwhile, excessive optimization in a single attribute may lead to a large weight of the attribute, causing the generated UAEs to deviate from the sampled images with unnatural semantic changes.
Additionally, an intriguing finding is that some neutral attributes such as \textit{smiling} tend to perform better than gender-related attributes like \textit{strong jawline} and \textit{busy eyebrows} when attacking the gender classification task.
We speculate that the male facial images in the dataset may appear more serious and lack smiles compared to female facial images, thus a negative weight of the \textit{smiling} attribute tends to lead the UAEs to be misclassified as a male.

\begin{table}[tbp]
	\caption{Ablation study of SemDiff attack against ResNet50. Weight denotes the mean absolute weight of attributes when the attack succeeds.} \label{tab:ablation}
	\centering
    \resizebox{\columnwidth}{!}{ 
	\begin{tabular}{c|ccccc}
        \hline
        {Method} & {ASR↑} & {BRISQUE↓} & {FID↓} & {KID↓} & {Weight↓}\\
        \hline
        {adv$\_$attr $\&$ multi} &\textbf{100\%} & 26.59 & \textbf{27.60} & \textbf{0.007} & \textbf{0.174}\\
        {attr $\&$ multi} & 76.6\% & \textbf{24.03} & 36.81 & 0.014 & 0.414\\
        \hline
        {adv$\_$strong$\_$jawline $\&$ single} & 99.2\% & 26.67 & 29.59 & 0.009 & 0.503\\
        {strong$\_$jawline $\&$ single} & 45.6\% & 28.51 & 33.48 & 0.012 & 1.270\\
        \hline
        {adv$\_$busy$\_$eyebrows $\&$ single} & 99.9\% & 31.71 & 31.80 & 0.010 &0.729\\
        {busy$\_$eyebrows $\&$ single} &69.3\% & 29.69 & 47.76 & 0.024 & 1.442\\
        \hline
        {adv$\_$smiling $\&$ single} & \textbf{100\%} & 25.24 & 28.86 & 0.008 & 0.433\\
        {smiling $\&$ single} & 90.6\% & 27.58 & 40.92 & 0.022 & 1.293\\
        \hline
	\end{tabular}
    }
\end{table}

\begin{figure}[tbp]
    \centering
    \includegraphics[width=\columnwidth]{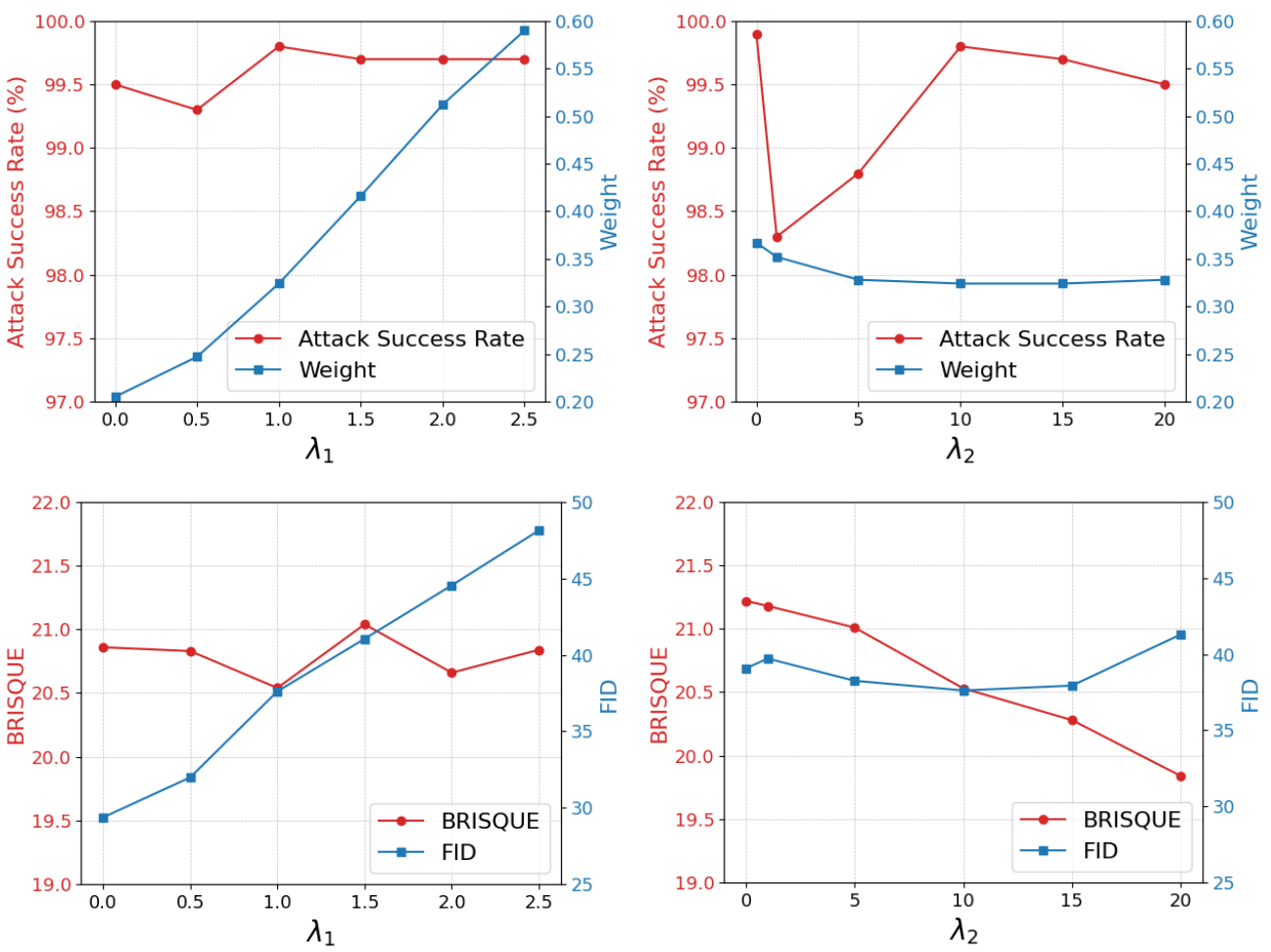}
    \caption{Hyperparameter analysis of $\lambda_{1}$ and $\lambda_{2}$ in SemDiff. The results are generated on CelebA-HQ dataset against ResNet50 model. We adopt the ASR, Weight, BRISQUE, FID and KID to measure the impact on attack effectiveness and generation quality.}
    \label{fig5}
\end{figure}

\subsection{Hyperparameter Analysis}
We analyze and discuss the impact of the important hyperparameters of SemDiff in the subsection.
Note that our proposed method does not require retraining the diffusion models, and the attack is performed in the reverse process.

\subsubsection{Influence of $\lambda_{1}$ and $\lambda_{2}$}
$\lambda_{1}$ and $\lambda_{2}$ are two hyperparameters used in the multi-attributes optimization approach.
$\lambda_{1}$ is the weight of reconstruction loss to keep UAEs resembling the source class $y_{source}$, and $\lambda_{2}$ is the weight of the penalty term to control the weight size and prevent excessive semantic shifts.
Thus, both $\lambda_{1}$ and $\lambda_{2}$ affect the balance between the attack effectiveness and the imperceptibility of the generated UAEs.
From Fig.~\ref{fig5}, we can observe that as $\lambda_{1}$ increases, the ASR and BRISQUE show small variation, while the Weight and FID also increase. 
This indicates a certain level of conflict between the objectives of causing the misclassification of target classifiers and maintaining the imperceptibility of UAEs to be consistent with $y_{source}$. 
Although achieving both goals requires larger attribute weights, $\lambda_{1}$ is indispensable to ensure the generated UAEs resemble the source class.
As for $\lambda_{2}$, as it increases, both the Weight and BRISQUE exhibit a gradual decline, which demonstrates that the weight penalty term effectively reduces semantic changes in the UAE, resulting in more natural UAEs close to the sampled images.
However, too large $\lambda_{2}$ will also influence the ASR and FID performance.
Thus, we set a moderate $\lambda_{1}$ and $\lambda_{2}$ to better balance the attack effectiveness and the imperceptibility of the generated UAEs.

\begin{figure*}[hbt]
    \centering
    \includegraphics[width=0.8\textwidth]{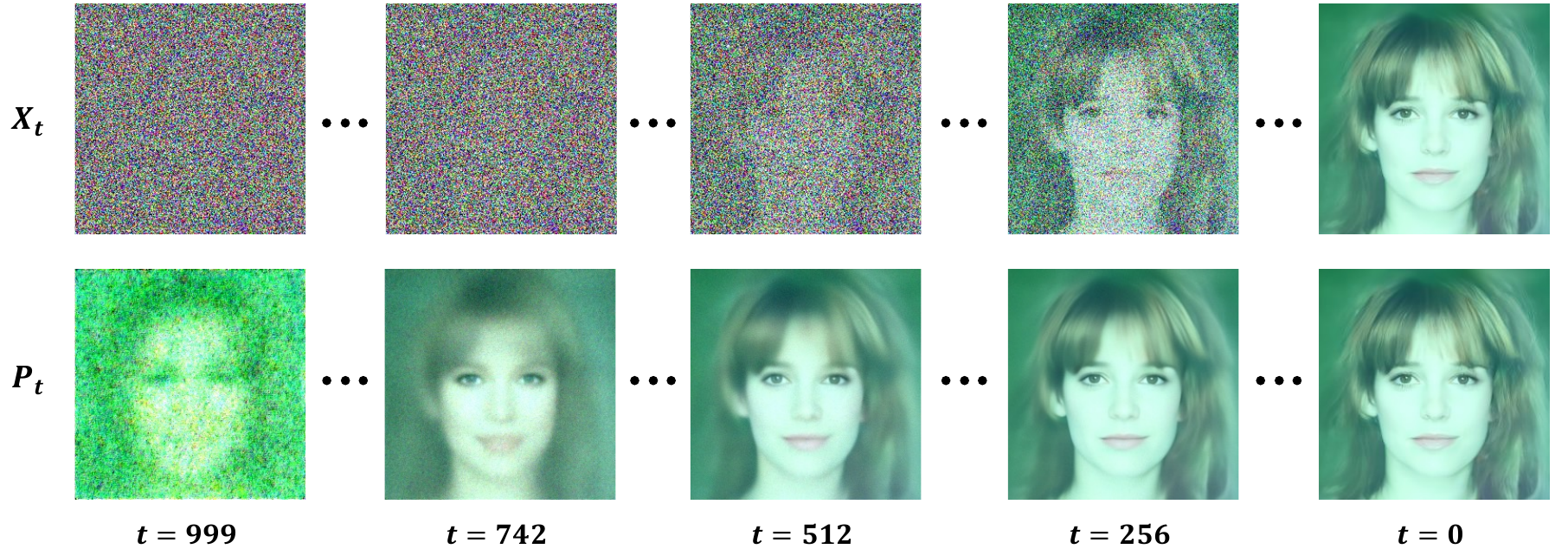}
    \caption{Visual examples of $\mathbf{P}_t$ or $\boldsymbol{x}_t$ at different timesteps in the denoising process of DDIM. The examples are generated by SemDiff on CelebA-HQ dataset.}
    \label{fig6}
\end{figure*}

\subsubsection{Choosing $\mathbf{P}_t$ or $\boldsymbol{x}_t$}
One main difference between our proposed SemDiff and previous diffusion-based attacks~\cite{17,18} is that we modify $\mathbf{P}_t$ and obtain the adversarial gradient through $\mathbf{P}_t$ instead of $\boldsymbol{x}_t$ in DDIM.
$\mathbf{P}_t$ is the predicted $\boldsymbol{x}_{0}$, and constitutes $\boldsymbol{x}_t$ together with $\mathbf{D}_t$ and random noise according to Eq. (\ref{eq6}).
Compared to $\boldsymbol{x}_t$, $\mathbf{P}_t$ better approximates the final output $\boldsymbol{x}_{0}$, and already contains clear semantic contents in the early stage of the denoising process, as shown in Fig.~\ref{fig6}.
To investigate the effect of choosing $\mathbf{P}_t$ or $\boldsymbol{x}_t$, we test our SemDiff equipped with $\mathbf{P}_t$ and $\boldsymbol{x}_t$, respectively.
From the results reported in Tab.~\ref{tab:hyperparameter}, we can observe that SemDiff with $\mathbf{P}_t$ outperforms SemDiff with $\boldsymbol{x}_t$ in all metrics, particularly achieving a significant lead in ASR.
This suggests that SemDiff with $\boldsymbol{x}_t$ can not obtain accurate adversarial gradients through noisy $\boldsymbol{x}_t$ in the early stage of the denoising process, leading to unsuccessful attacks and also affecting the naturalness and imperceptibility of the generated UAEs.
In comparison, the clear semantic contents in $\mathbf{P}_t$ in the early stage are enough to guide SemDiff to generate effective UAEs with adversarial semantic attribute shifts. 

\begin{table}[tbp]
	\caption{Ablation study of SemDiff attack against ResNet50. Weight denotes the mean absolute weight of attributes when the attack succeeds.} \label{tab:hyperparameter}
	\centering
    \resizebox{\columnwidth}{!}{ 
	\begin{tabular}{c|ccccc}
        \hline
        {Method} & {ASR↑} & {BRISQUE↓} & {FID↓} & {KID↓} & {Weight↓}\\
        \hline
        {SemDiff with $\mathbf{P}_t$} &\textbf{100\%} & \textbf{26.59} & \textbf{27.60} & \textbf{0.007} & \textbf{0.174}\\
        {SemDiff with $X_t$} & 65.7\% & 29.82 & 31.52 & 0.012 & 0.352\\
        \hline
	\end{tabular}
    }
\end{table}

\section{Conclusion}
\label{sec:conclusion}
In this paper, we present a novel unrestricted adversarial attack, SemDiff, to generate natural and imperceptible unrestricted adversarial examples (UAEs) with diffusion models.
Different from existing diffusion-based methods that simply optimize in the intermediate latent space, we first propose to utilize the semantic latent space of diffusion models to obtain adversarial but meaningful attributes for attack.
Specifically, we create a set of adversarial semantic attributes and train semantic functions to learn such attributes in order to guide the generation of diffusion models.
We further devise a multi-attributes optimization approach to optimize the weights of these attributes, so that the generated UAEs can achieve attack success while maintaining naturalness with semantic attributes changes.
Extensive experiments on four tasks on three high-resolution datasets, including CelebA-HQ, AFHQ, and ImageNet, demonstrate that SemDiff outperforms the state-of-the-art methods in both attack effectiveness and the imperceptibility of generated UAEs.
Moreover, we test SemDiff against various adversarial defenses, and the results show that SemDiff is capable of evading  different defenses, which further validates its threatening.
Considering the effectiveness and realism of the generated UAEs of SemDiff, we encourage researchers to further explore the vulnerability of deep neural networks and develop more robust methods to resist unrestricted adversarial attacks.


\bibliographystyle{IEEEtran}
\bibliography{Sections/references}

\begin{thebibliography}{10}
\providecommand{\url}[1]{#1}
\csname url@samestyle\endcsname
\providecommand{\newblock}{\relax}
\providecommand{\bibinfo}[2]{#2}
\providecommand{\BIBentrySTDinterwordspacing}{\spaceskip=0pt\relax}
\providecommand{\BIBentryALTinterwordstretchfactor}{4}
\providecommand{\BIBentryALTinterwordspacing}{\spaceskip=\fontdimen2\font plus
\BIBentryALTinterwordstretchfactor\fontdimen3\font minus
  \fontdimen4\font\relax}
\providecommand{\BIBforeignlanguage}[2]{{%
\expandafter\ifx\csname l@#1\endcsname\relax
\typeout{** WARNING: IEEEtran.bst: No hyphenation pattern has been}%
\typeout{** loaded for the language `#1'. Using the pattern for}%
\typeout{** the default language instead.}%
\else
\language=\csname l@#1\endcsname
\fi
#2}}
\providecommand{\BIBdecl}{\relax}
\BIBdecl

\bibitem{1}
J.~Deng, W.~Dong, R.~Socher, L.~Li, K.~Li, and L.~Fei{-}Fei, ``Imagenet: {A}
  large-scale hierarchical image database,'' in \emph{Proceedings of the 2009
  {IEEE} Computer Society Conference on Computer Vision and Pattern Recognition
  ({CVPR})}, Miami, FL, June 2009, pp. 248--255.

\bibitem{2}
O.~M. Parkhi, A.~Vedaldi, and A.~Zisserman, ``Deep face recognition,'' in
  \emph{Proceedings of the 26th British Machine Vision Conference ({BMVC})},
  Swansea, UK, September 2015, pp. 41.1--41.12.

\bibitem{3}
J.~Wu, W.~Fan, J.~Chen, S.~Liu, Q.~Li, and K.~Tang, ``Disentangled contrastive
  learning for social recommendation,'' in \emph{Proceedings of the 31st ACM
  International Conference on Information \& Knowledge Management ({CIKM})},
  2022, pp. 4570--4574.

\bibitem{4}
D.~Bahdanau, K.~Cho, and Y.~Bengio, ``Neural machine translation by jointly
  learning to align and translate,'' in \emph{Proceedings of the 3rd
  International Conference on Learning Representations ({ICLR})}, San Diego,
  CA, May 2015.

\bibitem{5}
C.~Szegedy, W.~Zaremba, I.~Sutskever, J.~Bruna, D.~Erhan, I.~J. Goodfellow, and
  R.~Fergus, ``Intriguing properties of neural networks,'' in \emph{Proceedings
  of the 2nd International Conference on Learning Representations ({ICLR})},
  Banff, AB, Canada, April 2014.

\bibitem{6}
I.~J. Goodfellow, J.~Shlens, and C.~Szegedy, ``Explaining and harnessing
  adversarial examples,'' in \emph{Proceedings of the 3rd International
  Conference on Learning Representations ({ICLR})}, San Diego, CA, May 2015.

\bibitem{7}
Z.~Dai, S.~Liu, Q.~Li, and K.~Tang, ``Saliency attack: towards imperceptible
  black-box adversarial attack,'' \emph{ACM Transactions on Intelligent Systems
  and Technology}, vol.~14, no.~3, pp. 1--20, 2023.

\bibitem{8}
L.-b. Ning, Z.~Dai, W.~Fan, J.~Su, C.~Pan, L.~Wang, and Q.~Li, ``Joint
  universal adversarial perturbations with interpretations,'' \emph{arXiv
  preprint arXiv:2408.01715}, 2024.

\bibitem{9}
L.-b. Ning, Z.~Dai, J.~Su, C.~Pan, L.~Wang, W.~Fan, and Q.~Li,
  ``Interpretation-empowered neural cleanse for backdoor attacks,'' in
  \emph{Companion Proceedings of the ACM on Web Conference ({WWW})}, Singapore,
  Singapore, May 2024, pp. 951--954.

\bibitem{10}
J.~Wu, N.~Lu, Z.~Dai, W.~Fan, S.~Liu, Q.~Li, and K.~Tang, ``Backdoor graph
  condensation,'' \emph{arXiv preprint arXiv:2407.11025}, 2024.

\bibitem{11}
S.~Liu, N.~Lu, C.~Chen, and K.~Tang, ``Efficient combinatorial optimization for
  word-level adversarial textual attack,'' \emph{IEEE/ACM Transactions on
  Audio, Speech, and Language Processing}, vol.~30, pp. 98--111, 2021.

\bibitem{12}
Y.~Song, R.~Shu, N.~Kushman, and S.~Ermon, ``Constructing unrestricted
  adversarial examples with generative models,'' in \emph{Annual Conference on
  Neural Information Processing Systems ({NeurIPS})}, Montr{\'{e}}al, Canada,
  December 2018, pp. 8322--8333.

\bibitem{13}
M.~Khoshpasand and A.~A. Ghorbani, ``On the generation of unrestricted
  adversarial examples,'' in \emph{50th Annual {IEEE/IFIP} International
  Conference on Dependable Systems and Networks Workshops}, Valencia, Spain,
  June 2020, pp. 9--15.

\bibitem{14}
L.~Zhang, N.~Yang, Y.~Sun, and P.~S. Yu, ``Provable unrestricted adversarial
  training without compromise with generalizability,'' \emph{{IEEE} Trans.
  Pattern Anal. Mach. Intell.}, vol.~46, no.~12, pp. 8302--8319, 2024.

\bibitem{15}
X.~Wang, K.~He, C.~Song, L.~Wang, and J.~E. Hopcroft, ``At-gan: An adversarial
  generator model for non-constrained adversarial examples,'' \emph{arXiv
  preprint arXiv:1904.07793}, 2019.

\bibitem{16}
T.~Xiang, H.~Liu, S.~Guo, Y.~Gan, and X.~Liao, ``{EGM:} an efficient generative
  model for unrestricted adversarial examples,'' \emph{{ACM} Trans. Sens.
  Networks}, vol.~18, no.~4, pp. 51:1--51:25, 2022.

\bibitem{17}
X.~Dai, K.~Liang, and B.~Xiao, ``Advdiff: Generating unrestricted adversarial
  examples using diffusion models,'' in \emph{European Conference on Computer
  Vision ({ECCV})}, A.~Leonardis, E.~Ricci, S.~Roth, O.~Russakovsky,
  T.~Sattler, and G.~Varol, Eds., vol. 15104, Milan, Italy, September 2024, pp.
  93--109.

\bibitem{18}
X.~Chen, X.~Gao, J.~Zhao, K.~Ye, and C.~Xu, ``Advdiffuser: Natural adversarial
  example synthesis with diffusion models,'' in \emph{{IEEE/CVF} International
  Conference on Computer Vision ({ICCV})}, Paris, France, October 2023, pp.
  4539--4549.

\bibitem{19}
A.~Huq and M.~T. Pervin, ``Analysis of adversarial attacks on skin cancer
  recognition,'' in \emph{International Conference on Data Science and Its
  Applications ({ICoDSA})}, 2020, pp. 1--4.

\bibitem{20}
P.~Dhariwal and A.~Q. Nichol, ``Diffusion models beat gans on image
  synthesis,'' in \emph{Annual Conference on Neural Information Processing
  Systems ({NeurIPS})}, M.~Ranzato, A.~Beygelzimer, Y.~N. Dauphin, P.~Liang,
  and J.~W. Vaughan, Eds., Virtual, December 2021, pp. 8780--8794.

\bibitem{21}
R.~Rombach, A.~Blattmann, D.~Lorenz, P.~Esser, and B.~Ommer, ``High-resolution
  image synthesis with latent diffusion models,'' in \emph{{IEEE/CVF}
  Conference on Computer Vision and Pattern Recognition ({CVPR})}, New Orleans,
  LA, June 2022, pp. 10\,674--10\,685.

\bibitem{22}
C.~Meng, Y.~He, Y.~Song, J.~Song, J.~Wu, J.~Zhu, and S.~Ermon, ``Sdedit: Guided
  image synthesis and editing with stochastic differential equations,'' in
  \emph{International Conference on Learning Representations ({ICLR})},
  Virtual, April 2022.

\bibitem{30}
J.~Ho, A.~Jain, and P.~Abbeel, ``Denoising diffusion probabilistic models,'' in
  \emph{Annual Conference on Neural Information Processing Systems
  ({NeurIPS})}, Virtual, December 2020.

\bibitem{23}
G.~Kim, T.~Kwon, and J.~C. Ye, ``Diffusionclip: Text-guided diffusion models
  for robust image manipulation,'' in \emph{{IEEE/CVF} Conference on Computer
  Vision and Pattern Recognition ({CVPR})}, New Orleans, LA, June 2022, pp.
  2416--2425.

\bibitem{24}
M.~Kwon, J.~Jeong, and Y.~Uh, ``Diffusion models already have {A} semantic
  latent space,'' in \emph{International Conference on Learning Representations
  ({ICLR})}, Kigali, Rwanda, May 2023.

\bibitem{25}
J.~Choi, J.~Lee, C.~Shin, S.~Kim, H.~Kim, and S.~Yoon, ``Perception prioritized
  training of diffusion models,'' in \emph{{IEEE/CVF} Conference on Computer
  Vision and Pattern Recognition ({CVPR})}, New Orleans, LA, June 2022, pp.
  2416--2425.

\bibitem{26}
A.~Madry, A.~Makelov, L.~Schmidt, D.~Tsipras, and A.~Vladu, ``Towards deep
  learning models resistant to adversarial attacks,'' in \emph{International
  Conference on Learning Representations ({ICLR})}, Vancouver, BC, Canada,
  April 2018.

\bibitem{27}
N.~Papernot, P.~D. McDaniel, S.~Jha, M.~Fredrikson, Z.~B. Celik, and A.~Swami,
  ``The limitations of deep learning in adversarial settings,'' in \emph{{IEEE}
  European Symposium on Security and Privacy ({EuroSP})}, Saarbr{\"{u}}cken,
  Germany, March 2016, pp. 372--387.

\bibitem{53}
L.~Huang, C.~Gao, and N.~Liu, ``Erosion attack: Harnessing corruption to
  improve adversarial examples,'' \emph{IEEE Transactions on Image Processing},
  vol.~32, pp. 4828--4841, 2023.

\bibitem{54}
D.~Wang, W.~Yao, T.~Jiang, and X.~Chen, ``Improving transferability of
  universal adversarial perturbation with feature disruption,'' \emph{IEEE
  Transactions on Image Processing}, vol.~33, pp. 722--737, 2023.

\bibitem{28}
A.~Odena, C.~Olah, and J.~Shlens, ``Conditional image synthesis with auxiliary
  classifier gans,'' in \emph{International Conference on Machine Learning
  ({ICML})}, vol.~70.\hskip 1em plus 0.5em minus 0.4em\relax Sydney, NSW,
  Australia: {PMLR}, August 2017, pp. 2642--2651.

\bibitem{29}
J.~Sohl{-}Dickstein, E.~A. Weiss, N.~Maheswaranathan, and S.~Ganguli, ``Deep
  unsupervised learning using nonequilibrium thermodynamics,'' in
  \emph{International Conference on Machine Learning ({ICML})}, vol.~37, Lille,
  France, July 2015, pp. 2256--2265.

\bibitem{31}
J.~Song, C.~Meng, and S.~Ermon, ``Denoising diffusion implicit models,'' in
  \emph{International Conference on Learning Representations ({ICLR})},
  Virtual, May 2021.

\bibitem{32}
K.~Preechakul, N.~Chatthee, S.~Wizadwongsa, and S.~Suwajanakorn, ``Diffusion
  autoencoders: Toward a meaningful and decodable representation,'' in
  \emph{{IEEE/CVF} Conference on Computer Vision and Pattern Recognition
  ({CVPR})}, New Orleans, LA, June 2022, pp. 10\,609--10\,619.

\bibitem{33}
Q.~Wu, Y.~Liu, H.~Zhao, A.~Kale, T.~Bui, T.~Yu, Z.~Lin, Y.~Zhang, and S.~Chang,
  ``Uncovering the disentanglement capability in text-to-image diffusion
  models,'' in \emph{{IEEE/CVF} Conference on Computer Vision and Pattern
  Recognition ({CVPR})}, Vancouver, BC, Canada, June 2023, pp. 1900--1910.

\bibitem{34}
R.~Rombach, A.~Blattmann, D.~Lorenz, P.~Esser, and B.~Ommer, ``High-resolution
  image synthesis with latent diffusion models,'' in \emph{{IEEE/CVF}
  Conference on Computer Vision and Pattern Recognition ({CVPR})}, New Orleans,
  LA, June 2022, pp. 10\,674--10\,685.

\bibitem{35}
R.~Gal, O.~Patashnik, H.~Maron, A.~H. Bermano, G.~Chechik, and D.~Cohen{-}Or,
  ``Stylegan-nada: Clip-guided domain adaptation of image generators,''
  \emph{ACM Transactions on Graphics ({TOG})}, vol.~41, no.~4, pp.
  141:1--141:13, 2022.

\bibitem{36}
A.~Radford, J.~W. Kim, C.~Hallacy, A.~Ramesh, G.~Goh, S.~Agarwal, G.~Sastry,
  A.~Askell, P.~Mishkin, J.~Clark, G.~Krueger, and I.~Sutskever, ``Learning
  transferable visual models from natural language supervision,'' in
  \emph{International Conference on Machine Learning ({ICML})}, M.~Meila and
  T.~Zhang, Eds., vol. 139.\hskip 1em plus 0.5em minus 0.4em\relax Virtual:
  {PMLR}, July 2021, pp. 8748--8763.

\bibitem{37}
T.~Karras, T.~Aila, S.~Laine, and J.~Lehtinen, ``Progressive growing of gans
  for improved quality, stability, and variation,'' in \emph{International
  Conference on Learning Representations ({ICLR})}, Vancouver, BC, Canada,
  April 2018.

\bibitem{38}
Y.~Choi, Y.~Uh, J.~Yoo, and J.~Ha, ``Stargan v2: Diverse image synthesis for
  multiple domains,'' in \emph{{IEEE/CVF} Conference on Computer Vision and
  Pattern Recognition ({CVPR})}, Seattle, WA, June 2020, pp. 8185--8194.

\bibitem{39}
K.~He, X.~Zhang, S.~Ren, and J.~Sun, ``Deep residual learning for image
  recognition,'' in \emph{{IEEE/CVF} Conference on Computer Vision and Pattern
  Recognition ({CVPR})}, Las Vegas, NV, June 2016, pp. 770--778.

\bibitem{40}
A.~Dosovitskiy, L.~Beyer, A.~Kolesnikov, D.~Weissenborn, X.~Zhai,
  T.~Unterthiner, M.~Dehghani, M.~Minderer, G.~Heigold, S.~Gelly, J.~Uszkoreit,
  and N.~Houlsby, ``An image is worth 16x16 words: Transformers for image
  recognition at scale,'' in \emph{International Conference on Learning
  Representations ({ICLR})}, Virtual, May 2021.

\bibitem{41}
K.~Simonyan and A.~Zisserman, ``Very deep convolutional networks for
  large-scale image recognition,'' in \emph{International Conference on
  Learning Representations ({ICLR})}, San Diego, CA, May 2015.

\bibitem{42}
T.~Karras, M.~Aittala, J.~Hellsten, S.~Laine, J.~Lehtinen, and T.~Aila,
  ``Training generative adversarial networks with limited data,'' in
  \emph{Annual Conference on Neural Information Processing Systems
  ({NeurIPS})}, Virtual, December 2020.

\bibitem{43}
A.~Brock, J.~Donahue, and K.~Simonyan, ``Large scale {GAN} training for high
  fidelity natural image synthesis,'' in \emph{International Conference on
  Learning Representations ({ICLR})}, New Orleans, LA, May 2019.

\bibitem{44}
M.~Heusel, H.~Ramsauer, T.~Unterthiner, B.~Nessler, and S.~Hochreiter, ``Gans
  trained by a two time-scale update rule converge to a local nash
  equilibrium,'' in \emph{Annual Conference on Neural Information Processing
  Systems ({NeurIPS})}, Long Beach, CA, December 2017, pp. 6626--6637.

\bibitem{45}
M.~Binkowski, D.~J. Sutherland, M.~Arbel, and A.~Gretton, ``Demystifying {MMD}
  gans,'' in \emph{International Conference on Learning Representations
  ({ICLR})}, Vancouver, BC, Canada, April 2018.

\bibitem{46}
A.~Mittal, A.~K. Moorthy, and A.~C. Bovik, ``Blind/referenceless image spatial
  quality evaluator,'' in \emph{Asilomar Conference on Signals, Systems and
  Computers ({ACSCC})}, M.~B. Matthews, Ed., Pacific Grove, CA, November 2011,
  pp. 723--727.

\bibitem{47}
D.~Hendrycks and T.~G. Dietterich, ``Benchmarking neural network robustness to
  common corruptions and perturbations,'' in \emph{International Conference on
  Learning Representations ({ICLR})}, May 2019.

\bibitem{48}
Z.~Liu, Q.~Liu, T.~Liu, N.~Xu, X.~Lin, Y.~Wang, and W.~Wen, ``Feature
  distillation: Dnn-oriented {JPEG} compression against adversarial examples,''
  in \emph{{IEEE/CVF} Conference on Computer Vision and Pattern Recognition
  ({CVPR})}, Long Beach, CA, June 2019, pp. 860--868.

\bibitem{49}
W.~Xu, D.~Evans, and Y.~Qi, ``Feature squeezing: Detecting adversarial examples
  in deep neural networks,'' in \emph{Annual Network and Distributed System
  Security Symposium ({NDSS})}, San Diego, California, February 2018.

\bibitem{50}
M.~Boroumand, M.~Chen, and J.~J. Fridrich, ``Deep residual network for
  steganalysis of digital images,'' \emph{{IEEE} Transactions on Information
  Forensics and Security}, vol.~14, no.~5, pp. 1181--1193, 2019.

\bibitem{51}
F.~Tram{\`{e}}r, A.~Kurakin, N.~Papernot, I.~J. Goodfellow, D.~Boneh, and P.~D.
  McDaniel, ``Ensemble adversarial training: Attacks and defenses,'' in
  \emph{International Conference on Learning Representations ({ICLR})},
  Vancouver, BC, Canada, April 2018, pp. 1--16.

\bibitem{52}
X.~Jia, Y.~Zhang, B.~Wu, J.~Wang, and X.~Cao, ``Boosting fast adversarial
  training with learnable adversarial initialization,'' \emph{IEEE Transactions
  on Image Processing}, vol.~31, pp. 4417--4430, 2022.

\end{thebibliography}


\begin{IEEEbiography}[{\includegraphics[width=1in,height=1.25in,clip,keepaspectratio]{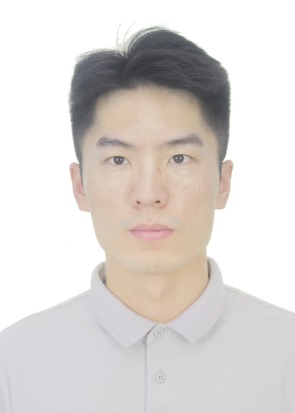}}]{Zeyu Dai}
    is currently a PhD candidate at The Hong Kong Polytechnic University (PolyU), Hong Kong, and Southern University of Science and Technology (SUSTech), Shenzhen, China. Previously, he received the B.Eng. degree in computer science and engineering in computer science and technology from Jiangnan University (JNU), Wuxi, China, in 2019. His research interests include adversarial machine learning and evolutionary computation. He has innovative works in top-tier conferences and journals.
\end{IEEEbiography}

\vspace{11pt}

\begin{IEEEbiography}[{\includegraphics[width=1in,height=1.25in,clip,keepaspectratio]{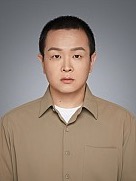}}]{Shengcai Liu}
    is currently a tenure-track assistant professor with the Department of Computer Science and Engineering, Southern University of Science and Technology (SUSTech), Shenzhen, China. He received Ph.D. degree in computer science and technology from the University of Science and Technology of China (USTC), Hefei, China, in 2020, respectively. His major research interests include learning to optimize and combinatorial optimization. He has authored or coauthored more than 25 papers in top-tier refereed international conferences and journals.
\end{IEEEbiography}

\vspace{11pt}

\begin{IEEEbiography}[{\includegraphics[width=1in,height=1.25in,clip,keepaspectratio]{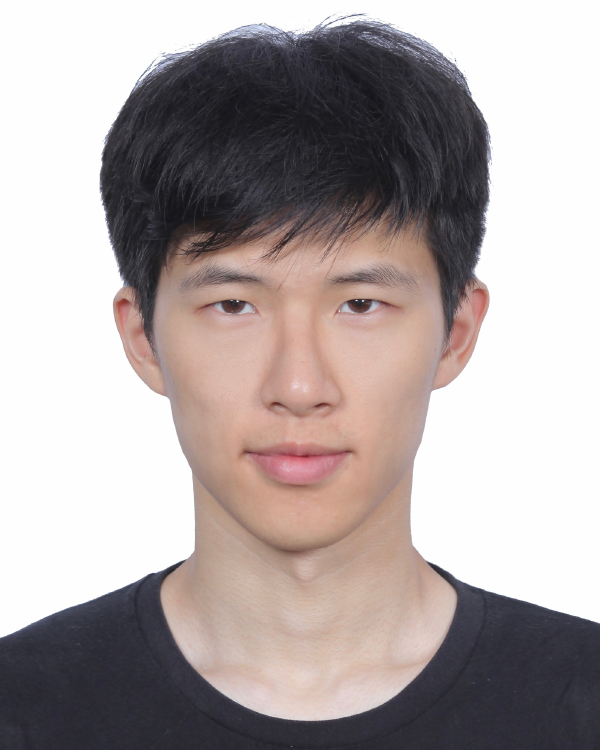}}]{Rui He}
    is currently a research engineer at BYD Auto. Previously, he obtained a BEng degree from the Southern University of Science and Technology (SUSTech), Shenzhen, China, in 2018 and a Ph.D. degree from the University of Birmingham (UoB) in 2024. His research interest focuses on active learning. He has innovative works in top conferences and journal.
\end{IEEEbiography}

\vspace{11pt}

\begin{IEEEbiography}[{\includegraphics[width=1in,height=1.25in,clip,keepaspectratio]{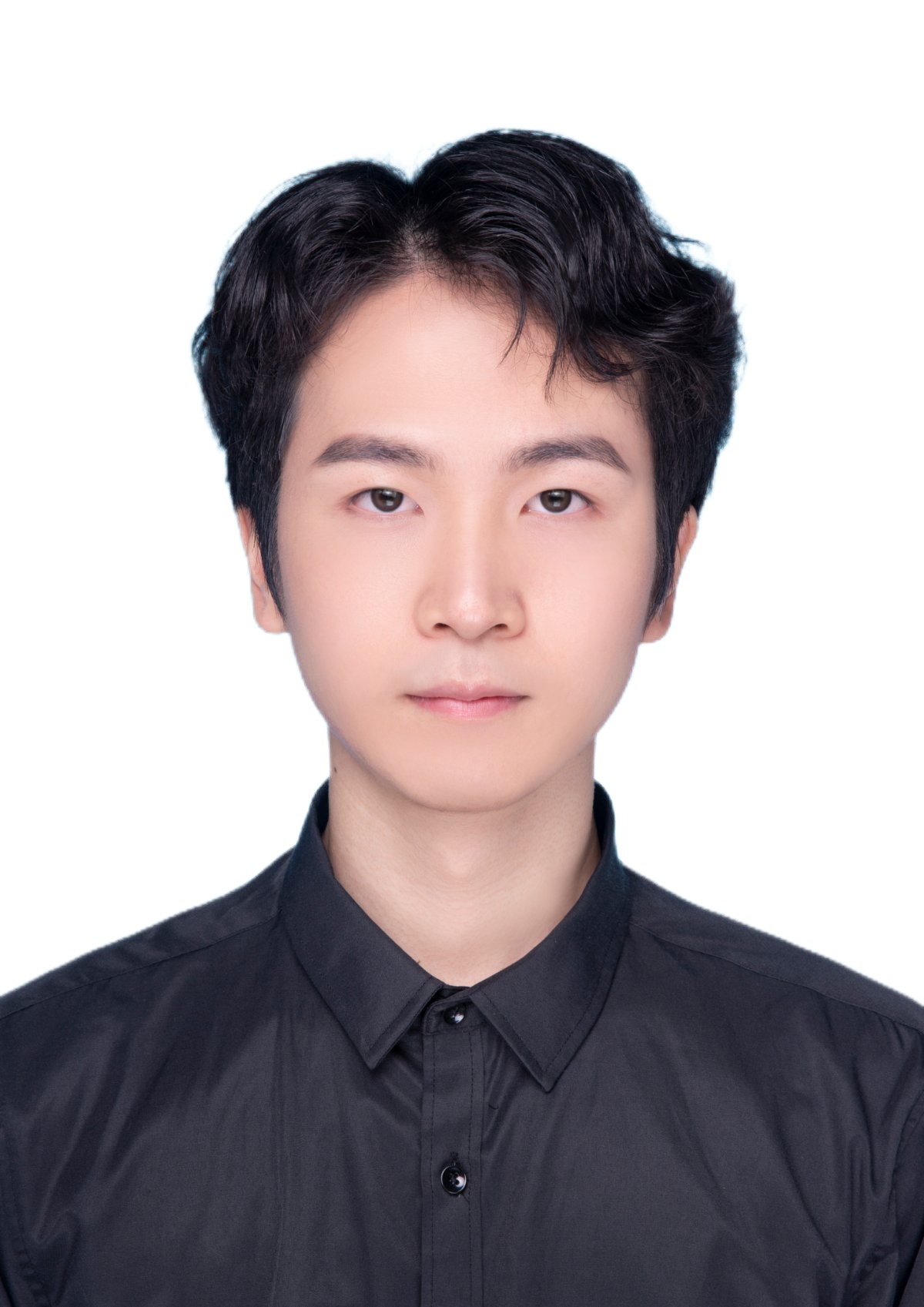}}]{Jiahao Wu}
    is currently a four-year Ph.D. candidate at The Hong Kong Polytechnic University (PolyU), Hong Kong, and Southern University of Science and Technology (SUSTech), Shenzhen, China. Previously, he earned his B.Eng. degree from the University of Science and Technology of China (USTC), Hefei, China, in 2020. His research interests include recommender systems and dataset condensation. He has innovative works in top-tier conferences.
\end{IEEEbiography}

\vspace{11pt}

\begin{IEEEbiography}[{\includegraphics[width=1in,height=1.25in,clip,keepaspectratio]{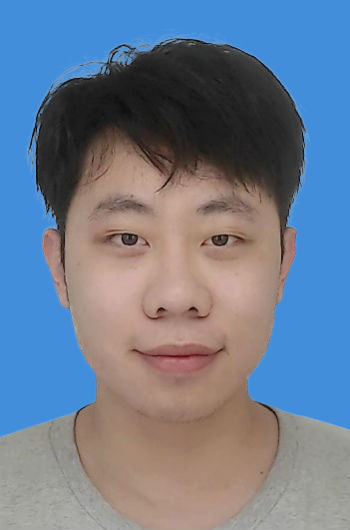}}]{Ning Lu}
    is currently working toward the Ph.D. degree of computer science in Hong Kong University of Science and Technology (HKUST) and Southern University of Science and Technology (SUSTech) from September 2021. Previously, he received the B.Eng. degree in computer science and engineering from the Southern University of Science and Technology, Shenzhen, China, in 2020. His current research topics include trustworthy AI and adversarial robustness of LLM. 
\end{IEEEbiography}

\vspace{11pt}

\begin{IEEEbiography}[{\includegraphics[width=1in,height=1.25in,clip,keepaspectratio]{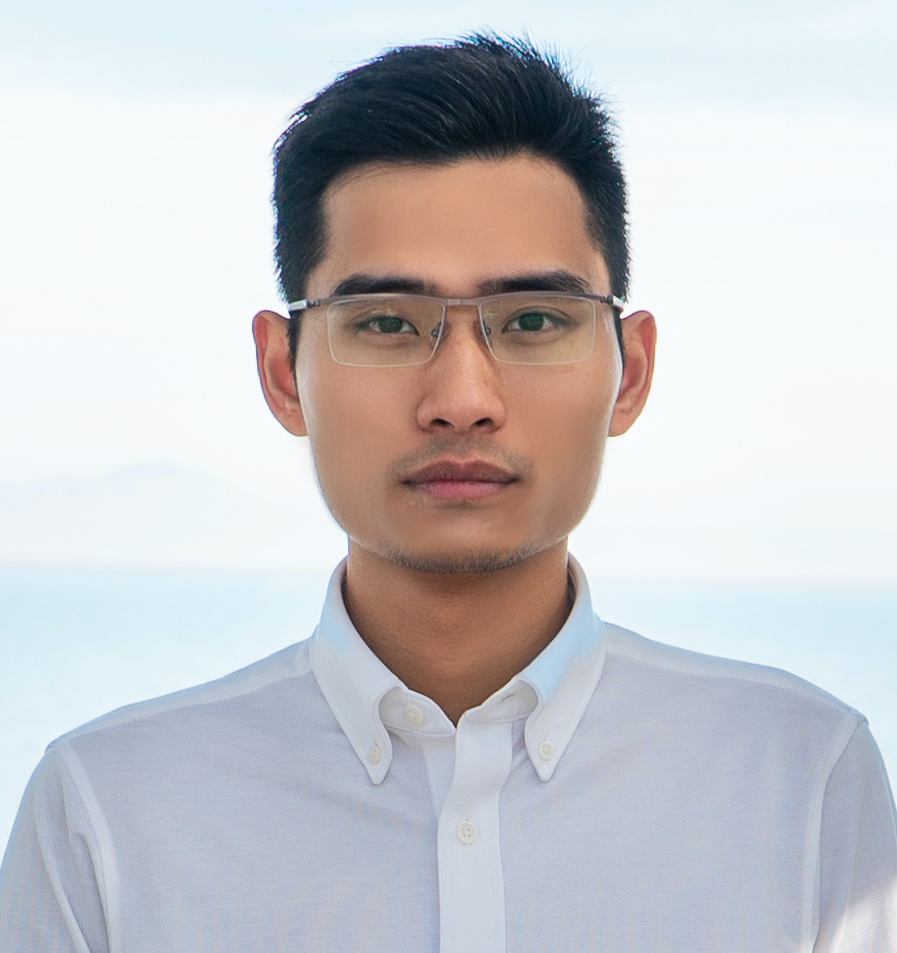}}]{Wenqi Fan}
    is currently an assistant professor of the Department of Computing at The Hong Kong Polytechnic University (PolyU), Hong Kong. He received his Ph.D. degree from the City University of Hong Kong (CityU), Hong Kong, in 2020. His research interests are in the broad areas of machine learning and data mining, with a particular focus on Recommender Systems, and Large Language Models. He has published innovative papers in top-tier journals and conferences. He serves as a top-tier conference (senior) PC member, session chair, and journal reviewer.
\end{IEEEbiography}

\vspace{11pt}

\begin{IEEEbiography}[{\includegraphics[width=1in,height=1.25in,clip,keepaspectratio]{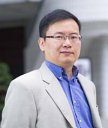}}]{Qing Li}
    is currently a Chair Professor (Data Science) and the Head of the Department of Computing, at the Hong Kong Polytechnic University (PolyU). He received a Ph.D. degree from the University of Southern California (USC), Los Angeles, USA. His research interests include object modeling, multimedia databases, social media, and recommender systems. He is a Fellow of IEEE and IET and a member of ACM SIGMOD and IEEE Technical Committee on Data Engineering. He has been actively involved in the research community by serving as an associate editor and reviewer for technical journals and as an organizer/co-organizer of numerous international conferences. He is the chairperson of the Hong Kong Web Society, and also served/is serving as an executive committee (EXCO) member of the IEEE-Hong Kong Computer Chapter and ACM Hong Kong Chapter. In addition, he serves as a councilor of the Database Society of Chinese Computer Federation (CCF), a member of the Big Data Expert Committee of CCF, and is a Steering Committee member of DASFAA, ER, ICWL, UMEDIA, and WISE Society.
\end{IEEEbiography}

\vspace{11pt}

\begin{IEEEbiography}[{\includegraphics[width=1in,height=1.25in,clip,keepaspectratio]{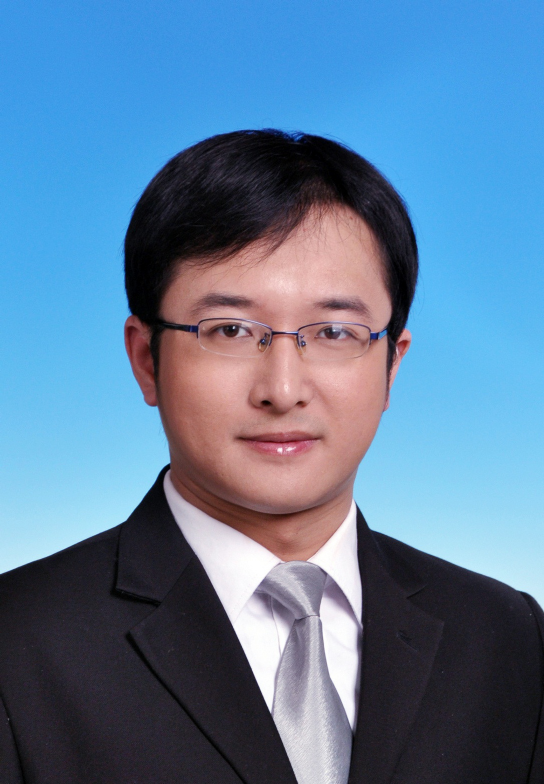}}]{Ke Tang}
    is currently a professor at the Department of Computer Science and Engineering, at Southern University of Science and Technology (SUSTech), Shenzhen, China. He received a Ph.D. degree in computer science from Nanyang Technological University (NTU), Singapore, in 2007. His research interests mainly include evolutionary computation, machine learning, and their applications. He is a Fellow of IEEE and he was awarded the Newton Advanced Fellowship (Royal Society) and the Changjiang Professorship Ministry of Education (MOE) of China. He was the recipient of the IEEE Computational Intelligence Society Outstanding Early Career Award and the Natural Science Award of MOE of China. He has been actively involved in the research community by serving as an organizer/coorganizer of numerous conferences, an associate editor of the IEEE TRANSACTIONS ON EVOLUTIONARY COMPUTATION, and as a member of editorial boards for a few other journals.
\end{IEEEbiography}


\vfill

\end{document}